\documentclass[10pt,journal,compsoc]{IEEEtran}
\PassOptionsToPackage{table,xcdraw}{xcolor}

\usepackage{amsmath,amsfonts}
\usepackage{array}
\usepackage{textcomp}
\usepackage{stfloats}
\usepackage{url}
\usepackage{verbatim}
\usepackage{graphicx}
\usepackage{booktabs}
\usepackage{cite}
\usepackage{color}
\usepackage[table]{xcolor}
\usepackage{bm}
\usepackage{multirow}
\usepackage[pagebackref,breaklinks,colorlinks]{hyperref}

\usepackage{cases}
\usepackage{booktabs}
\usepackage{graphicx}
\usepackage[ruled]{algorithm2e}
\usepackage{amssymb}
\usepackage{pifont}
\usepackage[table,xcdraw]{xcolor}
\usepackage{flushend}
\usepackage[normalem]{ulem}
\definecolor{mygray}{gray}{.9}

\usepackage{epsfig,latexsym}
\usepackage{float}
\usepackage{indentfirst}
\usepackage{amsmath}
\usepackage{times}
\usepackage{psfrag}
\usepackage{subfigure}
\usepackage{textcomp}
\usepackage{multirow}
\usepackage{multicol}
\usepackage{stfloats}
\usepackage{url}
\usepackage{subfigure}
\usepackage{booktabs}
\usepackage{comment}
\usepackage{threeparttable}
\usepackage{makecell}
\usepackage{bbding}

\definecolor{d-green}{rgb}{0.13, 0.55, 0.13}
\definecolor{d-red}{rgb}{0.55, 0.0, 0.0}

\PassOptionsToPackage{nocompress}{cite}
\ifCLASSOPTIONcompsoc
  \usepackage{cite}
\else
  \usepackage{cite}
\fi

\begin{document}

\title{Match Stereo Videos via Bidirectional Alignment}

\author{Junpeng Jing, Ye Mao, Anlan Qiu, Krystian Mikolajczyk
\IEEEcompsocitemizethanks{
\IEEEcompsocthanksitem  J. Jing, Y. Mao, A. Qiu, and K. Mikolajczyk are with the Department of Electrical and Electronic Engineering, Imperial College London, London SW7 2AZ, United Kingdom E-mail: \{j.jing23; ye.mao21; aq121; k.mikolajczyk\}@imperial.ac.uk. }}

\markboth{Journal of \LaTeX\ Class Files,~Vol.~14, No.~8, August~2015}%
{Shell \MakeLowercase{\textit{et al.}}: Bare Advanced Demo of IEEEtran.cls for IEEE Computer Society Journals}

\IEEEtitleabstractindextext{
\begin{abstract}
\textcolor{black}{Video stereo matching is the task of estimating consistent disparity maps from rectified stereo videos. There is considerable scope for improvement in both datasets and methods within this area. Recent learning-based methods often focus on optimizing performance for independent stereo pairs, leading to temporal inconsistencies in videos. Existing video methods typically employ sliding window operation over time dimension, which can result in low-frequency oscillations corresponding to the window size. To address these challenges, we propose a bidirectional alignment mechanism for adjacent frames as a fundamental operation. Building on this, we introduce a novel video processing framework, BiDAStereo, and a plugin stabilizer network, BiDAStabilizer, compatible with general image-based methods. Regarding datasets, current synthetic object-based and indoor datasets are commonly used for training and benchmarking, with a lack of outdoor nature scenarios. To bridge this gap, we present a realistic synthetic dataset and benchmark focused on natural scenes, along with a real-world dataset captured by a stereo camera in diverse urban scenes for qualitative evaluation.  Extensive experiments on in-domain, out-of-domain, and robustness evaluation demonstrate the contribution of our methods and datasets, showcasing improvements in prediction quality and achieving state-of-the-art results on various commonly used benchmarks. The project page, demos, code, and datasets are available at: \url{https://tomtomtommi.github.io/BiDAVideo/}.}
\end{abstract}

}

\maketitle

\IEEEdisplaynontitleabstractindextext

\IEEEpeerreviewmaketitle

\ifCLASSOPTIONcompsoc
\IEEEraisesectionheading{\section{Introduction}\label{sec:introduction}}
\else
\section{Introduction}
\label{sec:introduction}
\fi

\begin{figure*}[t]
  \centering
  \includegraphics[width=1\textwidth]{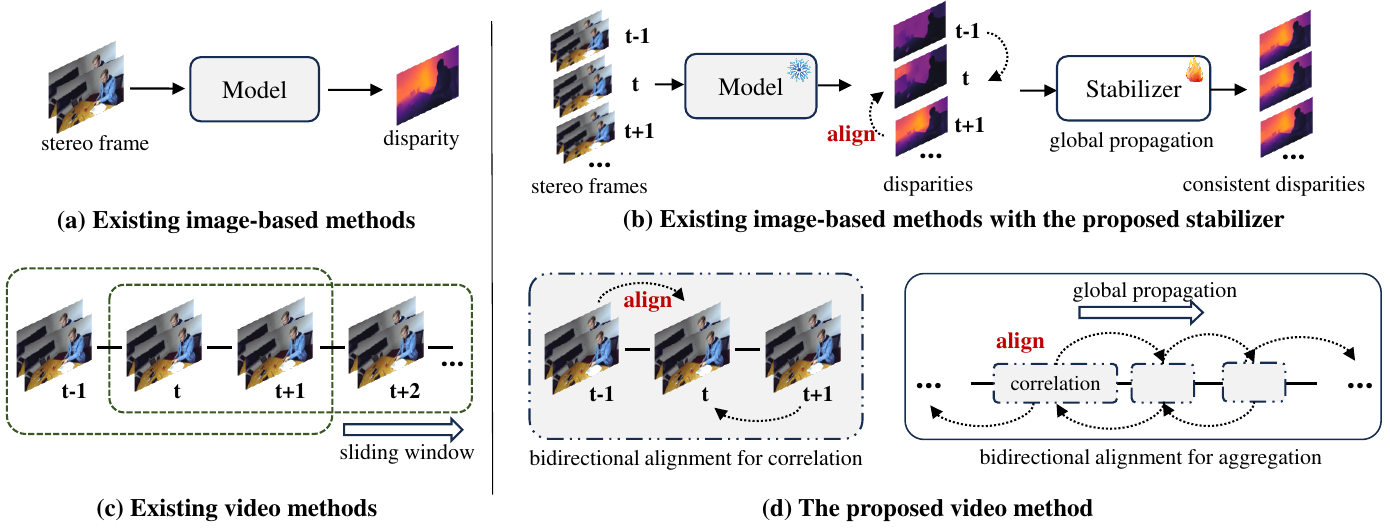}
  \vspace{-1.0em}
  \caption{Overview of the proposed methods. The illustration compares existing methods (left) with the proposed approaches (right). \textbf{(a)} Traditional image-based methods estimate disparity independently for each frame, leading to temporal inconsistency. \textbf{(b)} In contrast, our plugin stabilizer network takes inconsistent disparities as input to enhance temporal consistency without retraining the original stereo model. \textbf{(c)} Existing video methods segment sequences into fixed intervals, limiting information propagation to a short time frame range. \textbf{(d)} Our video method constructs cost volumes across neighboring frames and incorporates a recurrent update mechanism based on bidirectional alignment, ensuring global consistency throughout the entire sequence.}
  \label{fig:figure1}
\end{figure*}

\IEEEPARstart{S}{tereo} \textcolor{black}{matching is a fundamental task in computer vision, focusing on estimating the disparity between two rectified stereo images \cite{scharstein2002taxonomy}. This task is crucial in various applications, including 3D reconstruction \cite{geiger2011stereoscan}, robot navigation \cite{desouza2002vision}, and augmented reality (AR) \cite{azuma1997survey}. As consumer devices like AR glasses and smartphones with multiple cameras become more common, the need for accurate stereo matching is growing rapidly.}

\textcolor{black}{Deep learning has significantly advanced stereo-matching methods for estimating disparity, enhancing their accuracy \cite{lipson2021raft,li2022practical,xu2023iterative}, efficiency \cite{song2021adastereo,xu2022acvnet}, and robustness \cite{shen2021cfnet, Jing_2023_ICCV}. However, achieving consistent disparity estimations from stereo video sequences remains a challenge. When applying image-based methods directly to video sequences, severe flickering in disparity maps often occurs, as the processing is done on a frame-by-frame basis without leveraging temporal information, as illustrated in Fig.~\ref{fig:figure1}~(a). This issue is exacerbated in dynamic scenes where objects are in motion or deform. In such cases, multi-view constraints \cite{hartley2003multiple} are not applicable, and disparities lack translational invariance. Moreover, even when correspondences between frames are established, simply fusing independent disparity maps from corresponding points proves ineffective \cite{li2023temporally}.}

\textcolor{black}{Some recent approaches focused on leveraging information in neighbor frames for video stereo matching. Li \textit{et al.}\cite{li2023temporally} first proposed a general pipeline called CODD, which breaks down the problem into sub-modules for processing. This pipeline consists of a matching network for image-based disparity estimation, a motion network for SE3 transformation prediction, and a third network for temporal information fusion. However, this method is limited by its reliance on temporal information from only one past frame. To expand the temporal receptive field, DynamicStereo \cite{karaev2023dynamicstereo} was developed with self- and cross-attention mechanisms within a transformer architecture to extract and pool information across multiple frames. \textcolor{black}{Although extending the image-based version to a temporal one improves temporal consistency compared with purely image-based methods, its sliding-window approach still confines temporal information to a local range, as shown in Fig.~\ref{fig:figure1} (c). This limitation prevents the incorporation of global sequence information, leading to low-frequency oscillations at the scale of the temporal window size.} More importantly, since the positions of matching point pairs in stereo videos shift over time, directly computing temporal attention between raw video frames is often inaccurate. This raises a question: \textit{Is it possible to devise a mechanism that inherently facilitates feature matching in a video sequence and can be deployed with other existing methods?}}

\textcolor{black}{Spatial alignment in video stereo matching provides a promising solution to this problem. In video settings, matching quality can be improved by searching for potential correspondence points across the temporal dimension \cite{karaev2023dynamicstereo}, under the assumption that the reference point remains unchanged. This makes frame alignment especially important for videos with moving cameras, as it ensures that matching points are correctly aligned across different time steps. In this study, inspired by bidirectional alignment, we introduce  BiDAStereo, a video framework that achieves consistent video stereo matching and BiDAStabilizer, a plugin network for existing stereo image methods. As shown in Fig.~\ref{fig:figure1} (b), BiDAStabilizer is a plugin network designed to transform inconsistent disparity estimations from existing image-based methods into temporally consistent ones. It aligns disparities towards the center frame to extract local temporal cues from neighboring frames. Moreover, to leverage information from the entire sequence, we introduced a global propagation mechanism by aligning and fusing bidirectional feature states from neighboring frames to update the state of the center frame.  This method promotes recurrent propagation of global consistency, expanding the temporal receptive field and allowing the model to utilize a broader range of temporal information. \textcolor{black}{In BiDAStereo, as shown in Fig.~\ref{fig:figure1} (d), we apply alignment locally in the correlation module to build cost volumes and globally in the aggregation module to propagate information across the entire sequence. By aligning and aggregating information from both past and future frames, BiDAStereo achieves temporally consistent disparity estimation while effectively leveraging complementary frame content. This offers a significant advantage, especially in dynamic scenes.}}

\textcolor{black}{In addition to algorithmic challenges, existing synthetic video datasets are limited in realism and diversity. Widely-used dataset SceneFlow \cite{mayer2016large}, for instance, features abstract scenes with objects moving at different depths, but the scenes hardly reflect real-world cases. Although recent efforts, such as \cite{karaev2023dynamicstereo}, have introduced datasets featuring moving human and animal characters, they are still confined to indoor environments. Moreover, there is a noticeable lack of real-world resources suitable for visualization or qualitative evaluation. To overcome these limitations, we  provide a new synthetic dataset focused on outdoor natural scenes, and a real-world dataset encompassing a range of diverse indoor and outdoor urban environments.}

\textcolor{black}{The main contributions of this paper are as follows:
\begin{itemize}
\item
The bidirectional alignment mechanism is developed as an effective operation for enforcing temporal consistency in video stereo vision.
\item
A video framework is introduced, incorporating a triple-frame correlation layer and a motion-propagation recurrent unit, to utilize both local and global temporal information.
\item
A plugin stabilizer network is proposed to improve the temporal consistency of existing image-based stereo methods applied to video sequences.
\item
A synthetic stereo video benchmark is released, with high-quality annotations featuring natural scenes.
\item
A real-world stereo video dataset in diverse indoor and outdoor scenarios under different weather is introduced.
\item
The proposed methods achieve SOTA performance on various stereo video benchmarks.
\end{itemize}}

\textcolor{black}{\textit{Differences to conference version:} This work represents a substantial extension of our previous conference paper \cite{jing2024matchstereovideos}, with the key differences summarized as follows:
\begin{itemize}
\item
\textbf{Algorithm.} While the initial work introduced a video-based method, this paper extends it by making use of its bidirectional alignment technique in a lightweight stabilizer network that can be plugged into 
other existing methods. This allows to leverage the existing trained image-based methods for video tasks with improved results.
\item
\textbf{Dataset.} This paper introduces a realistic synthetic dataset for training and benchmarking methods on natural scenes, as well as a diverse real-world dataset captured by a stereo camera. These contributions bridge the gap in stereo video datasets, providing valuable resources for advancing the field.
\item 
\textbf{Experiment.} This work conducts extensive experiments with four additional metrics and two extra benchmarks compared to~\cite{jing2024matchstereovideos}. It encompasses in-domain, out-of-domain, and robustness evaluations from both quantitative and qualitative perspectives, providing a comprehensive comparative analysis.
\end{itemize}}

\textcolor{black}{The remainder of this paper is organized as follows. In Section 2, we review existing stereo matching methods as well as the current datasets available in this field. In Section 3, we introduce two novel datasets. Section 4 presents the detailed network architecture and training strategy. Section 5 presents the experimental results. Section 6 discusses the alignment method in the context of other related techniques. Section 7 concludes the paper.}

% \begin{table*}[tb] 
% \caption{Comparison of Stereo video datasets. }
% % \vspace{-.6em}
% \centering
% \setlength{\tabcolsep}{4.pt}
% \small
% \begin{tabular}{c|cccccccc}
% \rowcolor{mygray}
% \toprule[1.0pt]
% Dataset & Sintel  & SceneFlow & Falling & TartanAir & Dynamic & KITTI & Infinigen Stereo & South Kensington\\
% \rowcolor{mygray}
% Property &   \cite{} &  \cite{}  & Things \cite{} & \cite{} & Replica \cite{} & Depth \cite{} &  Video (ours) &  1024 (ours) \\
% \midrule
% Training frames  &   &  &  &  &  &  &   &   \\
% Test frames  &   &  &  &  &  &  &   &   \\
% Training sequences  &   &  &  &  &  &  &   &   \\
% Resolution  &   &  &  &  &  &  &   &   \\
% \midrule
% Disparity/Depth   &   &  &  &  &  &  &   &   \\
% \midrule
% Non-rigid objects   &   &  &  &  &  &  &   &   \\
% Realism  & synthetic  & synthetic & synthetic & synthetic & synthetic & real-world  & synthetic & real-world  \\
% Scenarios &   &  &  &  &  &  &   &   \\
% \bottomrule[1.0pt]
% \end{tabular}
% \label{tab:sintel-results}
% \end{table*}

\begin{table*}[t]
  \footnotesize
  \caption{Comparisons of stereo video datasets. Infinigen SV stands out for its focus on natural scenes, such as mountains and plains, which are not available at scale in previous datasets. SouthKen SV is distinguished by its manually collected real-world data in a diverse range of urban environments, captured using a stereo camera, both indoor and outdoor.}
  \vspace{-.5em}
  \begin{center}
\resizebox{\textwidth}{!}{
      \begin{tabular}{l|ccccccccc}
        \toprule[1.0pt]
        \rowcolor{mygray}
         & \textbf{Training} & \textbf{Test}  &   & \textbf{Disparity/} & \textbf{Optical} & \textbf{Seg.} & \textbf{Dynamic} &  &  \\ 
        \rowcolor{mygray}
        \multirow{-2}{*}{\textbf{Dataset Property}} & \textbf{frames / sequences} & \textbf{frames} & \multirow{-2}{*}{\textbf{Resolution}} & \textbf{Depth} & \textbf{Flow} & \textbf{Mask} & \textbf{objects} & \multirow{-2}{*}{\textbf{Scenario}} & \multirow{-2}{*}{\textbf{Type}}\\ 
        \hline
        Sintel \cite{sintel}     & 1064 / 23 & 564  & 1024×436 & \Checkmark & \Checkmark  & \Checkmark  & \Checkmark & Movie & Synthetic  \\ 
        SceneFlow \cite{mayer2016large}   & 34801 / 2256 & 4248  & 960x540  & \Checkmark & \Checkmark  & \Checkmark  & \Checkmark & Object & Synthetic  \\ 
        Falling Things \cite{fallingthings}  & 60200 / 330 & 0  & 960x540 & \Checkmark  & \XSolidBrush  & \Checkmark & \XSolidBrush  & Object &  Synthetic \\ 
        TartanAir \cite{tartanair2020iros}   & 296000 / 1037 & 0 & 640x480 & \Checkmark  & \Checkmark  & \Checkmark & \XSolidBrush  & Various &  Synthetic \\ 
        XR-Stereo \cite{cheng2024stereo} & 60000 / 10 & 2 & 640x480 & \Checkmark  & \Checkmark  & \Checkmark & \XSolidBrush  & Indoor &  Synthetic \\ 
        Dynamic Replica \cite{karaev2023dynamicstereo}  & 145200 / 484 & 18000 & 1280x720 & \Checkmark  & \Checkmark  & \Checkmark & \Checkmark  & Indoor &  Synthetic \\ 
        KITTI Depth \cite{kittidepth}  & 44506 / 142 & 3426 & 1242x375 & (sparse)  & \XSolidBrush  & \XSolidBrush  & \Checkmark & Road & Real  \\ 
        Infinigen SV (ours)  & 16800 / 186   & 1848 & 1280x720 & \Checkmark  & \Checkmark  & \Checkmark & \Checkmark  & Nature &  Synthetic \\ 
        SouthKen SV (ours)   & - &107821 & 1280x720 & (pseudo)  & (pseudo)  & (pseudo) & \Checkmark  & Various &  Real \\ 
        \bottomrule[1.0pt]
      \end{tabular}
}
  \end{center}
  \label{tab:dataset}
\end{table*}

\section{Related Works}
% \textcolor{black}{In this section, we first review image-based stereo matching methods. Next, we discuss recent advancements in stereo video methods. Finally, we review stereo video datasets.}

\subsection{Image Stereo Matching}
\textcolor{black}{Stereo matching is a classic computer vision  problem extensively researched over several decades. Traditional approaches to stereo matching \cite{kanade1994stereo, hsieh1992performance, scharstein1998stereo} can be categorized into local and global methods. Local methods divide images into patches, calculate matching costs between these patches, and subsequently perform local aggregation \cite{birchfield1999depth, van2002hierarchical}. In contrast, global methods formulate the task as an energy optimization problem, employing an explicit cost function that is optimized by belief propagation or graph cut algorithms \cite{sun2003stereo, klaus2006segment, boykov2001fast}.}

\textcolor{black}{Zbontar and LeCun \cite{zbontar2015computing} were the first to propose convolutions for computation of the matching cost. Mayer \textit{et al.}~\cite{mayer2016large} further advanced this by introducing an end-to-end network. Subsequent research on end-to-end networks can be categorized into two types: correlation-based methods with either 3D or 4D cost volumes. In a 3D cost volume, the dimensions are height, width, and disparity, while a 4D cost volume includes an additional feature dimension. The first category usually utilizes 2D convolutions for cost aggregation, which is efficient due to the decimation of feature channels. In this direction, innovative mechanisms and modules have been proposed, such as cascaded connection in CRL \cite{pang2017cascade}, group-wise correlation in GWCNet \cite{guo2019group}, adaptive aggregation in AANet \cite{xu2020aanet}, and hierarchical connection in HITNet \cite{tankovich2021hitnet}. Alternatively, another direction focuses on 3D convolutions for cost aggregation, which has better performance in terms of accuracy. Various architectural designs and modules have been investigated, such as the 3D hourglass aggregation module in GCNet \cite{kendall2017end}, spatial pyramid connection in PSMNet \cite{chang2018pyramid}, semi-global guided aggregation in GANet \cite{zhang2019ga}, and high-resolution targeted multi-scale layer in HSMNet \cite{yang2019hierarchical}. More recently, an iterative mechanism has demonstrated its effectiveness \cite{teed2020raft}. RAFTStereo \cite{lipson2021raft} first introduced this approach to stereo matching, achieving significant improvements. Building on this, CREStereo \cite{li2022practical} proposed adaptive correlation and incorporated attention mechanisms, yielding performance enhancements. IGEVStereo \cite{xu2023iterative} combined a geometry encoding volume with the backbone of RAFTStereo and used the results from the 3D aggregation network as the initial input to the iterative refinement.}

\textcolor{black}{In addition to accuracy and efficiency, some methods focus on robustness \cite{pang2018zoom, zhang2022revisiting, chuah2022itsa, shen2021cfnet, ucfnet, chang2023domain, rao2023masked, Jing_2023_ICCV}, which is crucial in real-world scenarios. Pang \textit{et al.} \cite{pang2018zoom} proposed a self-adaptation method to generalize to a target domain without ground truth. Song \textit{et al.} \cite{song2021adastereo} proposed a domain adaptation approach to narrow the gap between synthetic and real-world domains. CFNet and UCFNet \cite{shen2021cfnet, ucfnet} developed a cascade and fused cost volume representation and variance-based uncertainty estimation to reduce domain differences. Rao \textit{et al.} \cite{rao2023masked} introduced masked representation learning for domain-generalized stereo matching. Chang \textit{et al.} \cite{chang2023domain} devised a hierarchical visual transformation network to diversify the distribution of training data. Jing \textit{et al.} \cite{Jing_2023_ICCV} introduced an uncertainty-guided adaptive warping layer for different disparity distributions and domain gaps. Despite these notable strides, these methods primarily address image-based stereo matching, neglecting temporal information in video sequences. Direct application to video frames often results in poor temporal consistency, notably manifested as severe flickering in the output disparities.}

\subsection{Video Stereo Matching}
\textcolor{black}{Matching stereo videos is the integration of two objectives: static stereopsis and temporal correspondence \cite{jenkin1986applying}. Monocular video depth \cite{luo2020consistent, kopf2021robust, li2023towards} is a closely related field. However, unlike stereo matching, which focuses on ``real'' depth, these methods address scale-ambiguity, where a 3D point’s depth may vary over time due to inconsistent depth scales.}

\textcolor{black}{Traditional approaches \cite{sand2004video, min2010temporally} assume shared planes among neighboring frames and use 3D spatial windows and temporal propagation to estimate disparities. In the era of deep learning, Zhong \textit{et al.}~\cite{zhong2018open} introduced an LSTM-based approach for stereo videos, focusing on unsupervised learning with limited real-world annotations but neglecting temporal information. For dynamic scenes, Li \textit{et al.}\cite{li2023temporally} proposed CODD, emphasizing temporal consistency. They expanded a image-based stereo network with separate motion and fusion networks to align and aggregate current and past estimated disparities. Zhang \textit{et al.}\cite{zhang2023temporalstereo} developed a coarse-to-fine network leveraging past context to enhance predictions in challenging scenarios like occlusions and reflective regions. \textcolor{black}{Cheng \textit{et al.}\cite{cheng2024stereo} contributed by creating a synthetic dataset for indoor XR scenarios. They also introduced a model called XRStereo, which requires extra camera pose information for temporal warping between frames for temporal consistency.} Khan \textit{et al.}\cite{khan2023temporally} proposed a learned fusion approach by updating the global point cloud of each frame. However, these methods are limited to propagating temporal information only within the past neighboring frame, thereby constraining the temporal receptive field. \textcolor{black}{They also heavily rely on auxiliary known camera motions or scene geometry priors for fusing temporal information, limiting the choice of training datasets and practical application.} Karaev \textit{et al.}~\cite{karaev2023dynamicstereo} expanded the receptive field via a transformer-based architecture with time, stereo, and temporal attention mechanisms. \textcolor{black}{More recently, Jing \textit{et al.}~\cite{jing2025stereovideotemporallyconsistent} adapted monocular priors to enhance the feature representation.} Despite achieving improved results through sliding window processing, their approach confines temporal information within a fixed pre-set range, falling short of encompassing the entire sequence. Besides, the lack of alignment operations renders it suboptimal for cost aggregation.}

\begin{figure*}[t]
  \centering
  \includegraphics[width=1\textwidth]{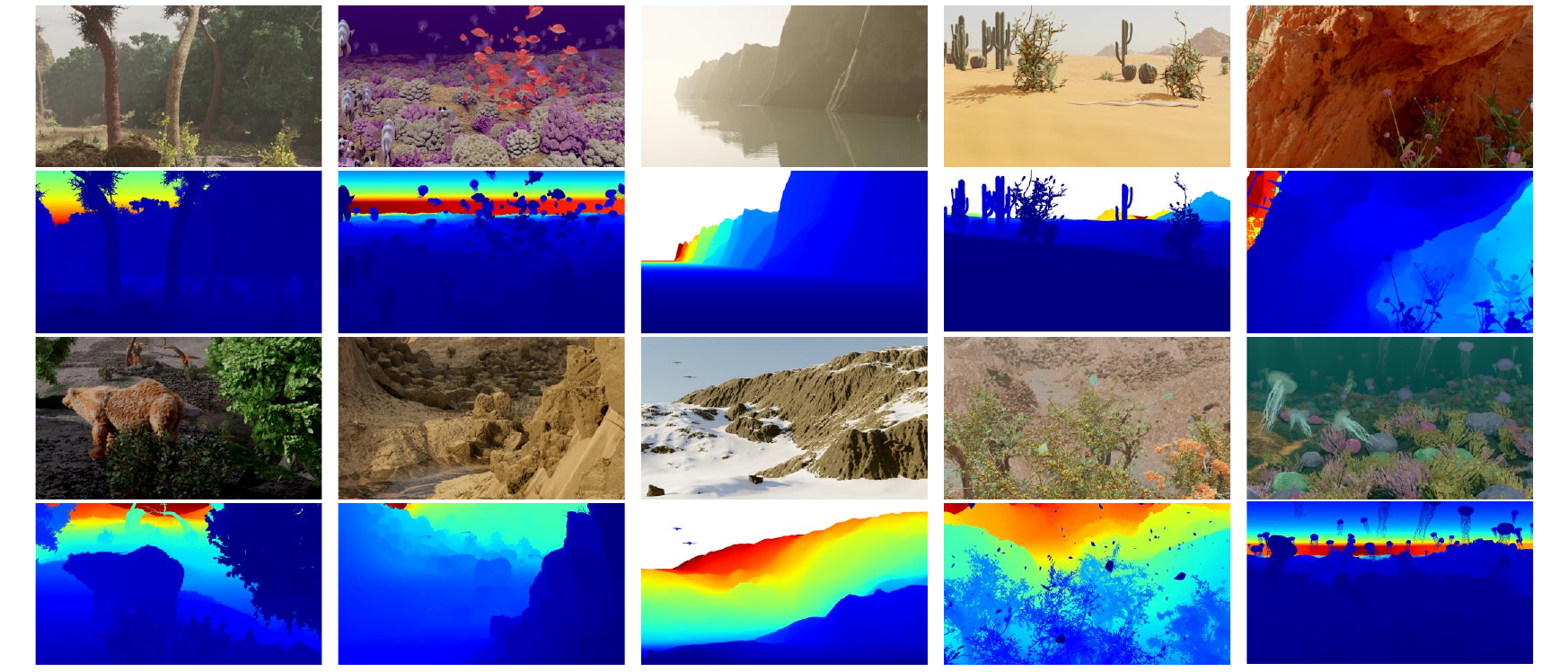}
  \vspace{-.5em}
  \caption{Infinigen Stereo Video (Infinigen SV) Dataset. Example video frames and depth maps from the proposed dataset. This dataset features a diverse range of outdoor nature scenes, offering a large-scale training set and a benchmark for disparity estimation in near-realistic environments.}
  % \vspace{-1.0em}
  \label{fig:infinigen}
\end{figure*}

\subsection{Stereo Video Datasets} 
\textcolor{black}{Training data is crucial for supervised learning-based methods in stereo matching. Collecting real-world data for this task poses significant challenges due to the difficulty in capturing ground truth depth and the even greater difficulty in obtaining consistent depth in video frames. As shown in Tab.~\ref{tab:dataset}, KITTI Depth \cite{kittidepth} is the only real-world stereo video dataset with ground truth depth derived from laser sensors. However, the process of ground truth generation is especially challenging considering the individually moving objects since they cannot be easily recovered from laser scanner data due to the rolling shutter of the Velodyne and the low framerate (10 fps). Different from KITTI 2015 \cite{kitti}, KITTI Depth does not separately optimize static background and dynamic objects or compensate for the vehicle’s ego-motion. Thus, strictly speaking, this dataset is suitable for monocular depth estimation, but should not be considered an accurate and consistent benchmark for stereo video. Additionally, the scenarios are limited to driving scenes and the ground truth annotation is sparse.}

\textcolor{black}{Synthetic datasets offer a promising alternative by providing accurate and dense ground truth. Sintel \cite{sintel}, the first dataset for disparity and optical flow estimation, is limited in scale and thus unsuitable for comprehensive training. SceneFlow \cite{mayer2016large} offers a large-scale dataset and serves as a fundamental training resource for most learning-based stereo methods. However, the objects are randomly placed, and the texture and background are randomly projected, lacking realism. Moreover, the number of frames for most videos is within $10$, which limits the training process. Falling Things \cite{fallingthings} enhances background quality but does not include dynamic objects. TartanAir \cite{tartanair2020iros}, a SLAM dataset, contains few dynamic objects and has low resolution. Recently, Dynamic Replica \cite{karaev2023dynamicstereo} introduced a large-scale, semi-realistic synthetic indoor dataset focused on dynamic objects, addressing some of these limitations. Despite the effectiveness of these datasets, we still lack diverse and large-scale benchmarks for outdoor nature scenes. We also lack diverse high-quality video data in real-world scenes including indoor and outdoor environments.}

\begin{figure*}[t]
  \centering
  \includegraphics[width=1\textwidth]{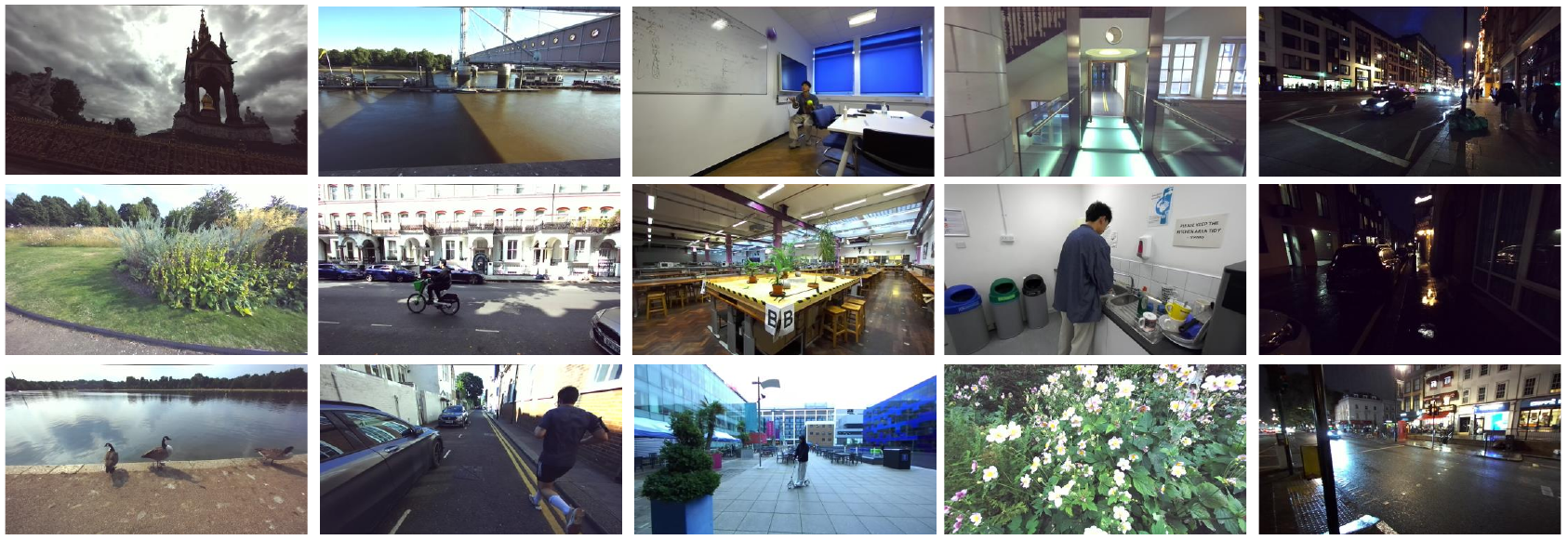}
  \vspace{-.5em}
  \caption{South Kensington Stereo Video (SouthKen SV) Dataset. The proposed dataset contains diverse indoor and outdoor scenes from the South Kensington area in London, UK, captured using a stereo camera, providing a qualitative benchmark for real-world disparity estimation.}
  % \vspace{-1.0em}
  \label{fig:southken}
\end{figure*}

% \begin{figure}[tbp]
%    \begin{center}
%    \includegraphics[width=.75\linewidth]{figures/camera.pdf}
%    \end{center}
%     \vspace{0em}
%       \caption{The dataset acquisition setup comprises a 3D-printed holder, a stereo camera, and a hand-held electronic stabilizer.} 
%    \label{fig:camera}
%    \vspace{0em}
% \end{figure}

\begin{figure}[tbp]
   \begin{center}
   \includegraphics[width=.88\linewidth]{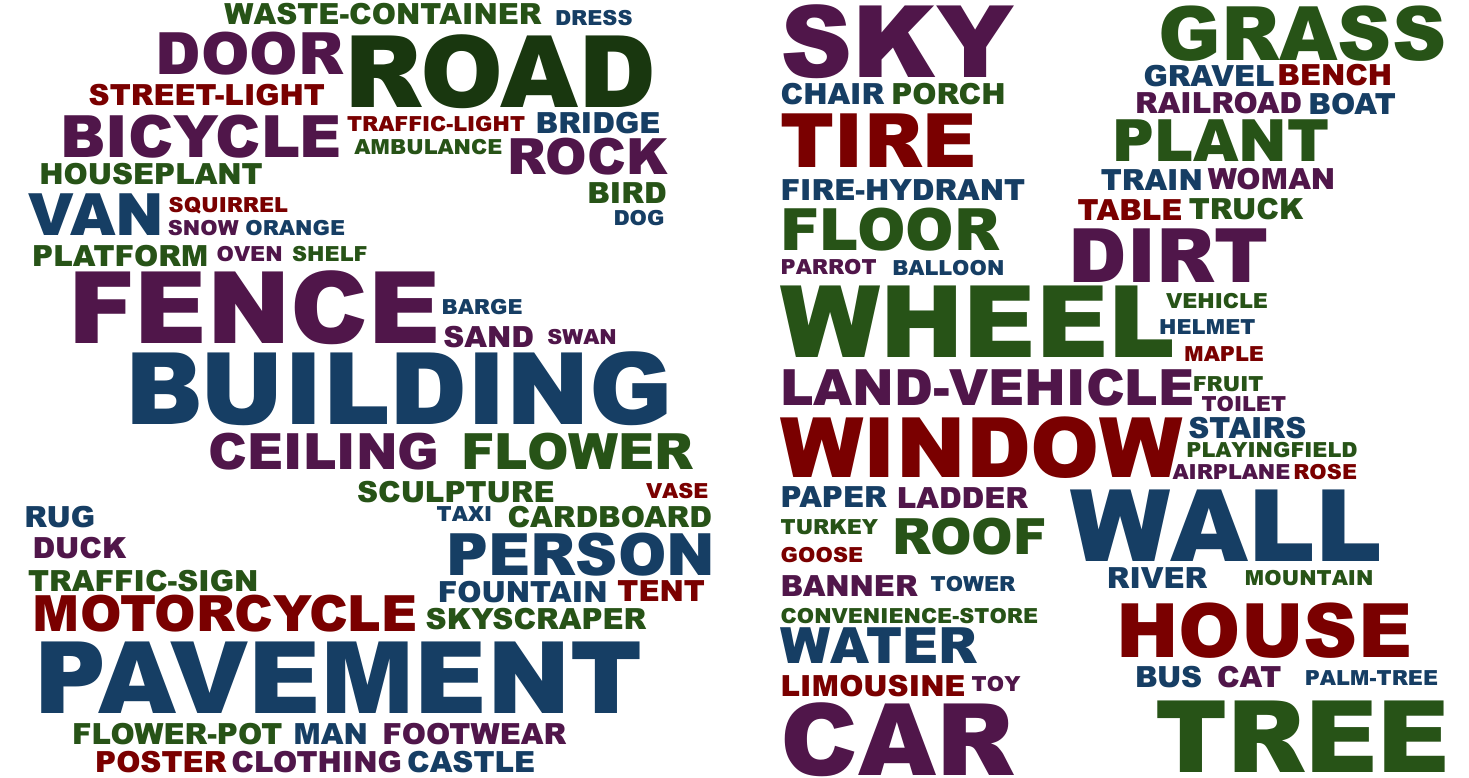}
   \end{center}
    \vspace{-.5em}
      \caption{Objects presented in SouthKen SV dataset.} 
   \label{fig:bubble}
\end{figure}

\section{Proposed Datasets}
% In this section, we introduce the proposed synthetic and real-world stereo video datasets.

\subsection{Infinigen Stereo Video Dataset}
\textcolor{black}{A high-quality synthetic stereo video dataset that offers dense ground-truth annotations is crucial for training temporally consistent stereo matching models. As shown in Tab.~\ref{tab:dataset}, while both SceneFlow \cite{mayer2016large} and Dynamic Replica \cite{karaev2023dynamicstereo} meet this requirement, SceneFlow is limited to very short sequences (mostly within 10 frames) featuring randomly textured moving objects, and Dynamic Replica focuses primarily on indoor scenes. Both datasets offer limited coverage of natural environments. For applications such as geological surveys, drone navigation, ecological monitoring, rescue robots, agricultural automation, and outdoor robotics in general, accurate perception of the natural environment is essential. To address this, we introduce \textit{Infinigen Stereo Video}, a diverse synthetic dataset and benchmark for outdoor nature scenes.}

\textcolor{black}{As detailed in Tab.~\ref{tab:dataset} and Fig.~\ref{fig:infinigen}, Infinigen SV dataset includes 226 videos of photorealistic virtual scenes, each lasting 3-20 seconds with a frame rate of 24 fps. The dataset is split into 186 training videos, 10 validation videos, and 30 testing videos. Following Dynamic Replica \cite{karaev2023dynamicstereo}, the videos are rendered at a resolution of $1280\times720$, higher than those in existing stereo datasets like Sintel ($1024\times536$) and SceneFlow ($960\times540$). For each scene, we generate a camera trajectory with smooth movements at a randomly sampled speed between 0.5 \textit{m/s} and 1.5 \textit{m/s}. This dataset is created using the free and open-source platform Infinigen \cite{raistrick2023infinite}, which is based on Blender \cite{blender}. It features a wide variety of 3D assets with detailed geometry and textures, including rocks, plants, animals, and natural phenomena such as clouds, rain, and snow. The dataset covers various natural terrains, including arctic, canyon, cave, cliff, coast, desert, forest, mountain, plain, river, and underwater scenes. To ensure broad generalization across different stereo setups, camera baselines are randomized within a range of 75 \textit{mm} to 1205 \textit{mm}.}

\textcolor{black}{Unlike other synthetic datasets that employ real-time rendering engines like Lumen \cite{skorobogatova2022real} or Eevee \cite{blender} for efficiency and low computational cost, we use Blender’s path tracing engine Cycles \cite{blender} to simulate complex real-world optics and render images in high fidelity. Each sample video includes ground-truth rendered depth maps, optical flows, surface normals, occlusion boundaries, instance segmentation masks, albedo, pixel-wise raw light intensity, specular reflection, and camera poses for both stereo views. This dataset also supports the development of generic computer vision tasks such as 3D reconstruction, monocular depth estimation, structure from motion, and SLAM. The overall pipeline was implemented using 8 NVIDIA Tesla A100 80GB GPUs, and rendering the entire dataset took approximately 6 months.}

\begin{table}[tbp] \addtolength{\tabcolsep}{-2pt}
\footnotesize
   \begin{center}
       \caption{Summary of the notations in this paper.}
      \vspace{-.5em}
     \begin{tabular}{c|l}
        \toprule[1.0pt]
        \rowcolor{mygray}
         \textbf{Notation} & \textbf{Description} \\
         \midrule
         $H$ / $W$ / $C$ & The height / weight / channel number of the feature \\
         $s$ & The downsampled scale \\
         $n$ / $N$ & The $n$-th / number of iterations \\
         $t$ / $T$ & The $t$-th / number of frames \\
         $\mathcal{A}(\cdot)$ & The alignment operation  \\
         $\mathbf{I}_{L}$ / $\mathbf{I}_{R}$ & The left / right RGB frame \\
         $\mathbf{f}_f$ / $\mathbf{f}_b$& The forward / backward optical flow \\
         $\mathbf{F}_{d}$ & The disparity feature map\\
         $\mathbf{F}_{f}$ / $\mathbf{F}_{b}$ & The forward / backward hidden state feature map\\
         $\mathbf{F}_L / \mathbf{F}_R$ & The left / right frame feature map \\
         $\mathbf{p}$ & The position of a pixel in correlation \\
         $\mathbf{r}$ / $\mathbf{R}$ & The delta position / search range in correlation\\
         $\mathbf{C}^{t\rightarrow t}$ & The correlation between the $t$-th left and right frames \\
         $\mathbf{M}^t$ &  The motion hidden state feature map\\
         $\hat{\mathbf{F}} / \hat{\mathbf{M}}$  & The aligned feature via optical flow \\
         $\tilde{\mathbf{F}}_{R}$ & The aligned right frame via disparity \\
         $\mathbf{F}_{(corr)}$ & The correlation feature map\\
         $\mathbf{F}_{(disp)}$ & The disparity feature map\\
         $\mathbf{F}_{(mot)}$ & The motion feature map\\
         $\mathbf{F}_{(prop)}$ & The propagation feature map \\
         $\mathbf{h}$ & The disparity hidden state \\
         $\left \langle \cdot \right \rangle $ & The channel-wise product operation \\
         $\mathbf{d}$ / $\mathbf{\Delta d}$ & The predicted disparity / residual disparity \\
         $\mathbf{O}$ & The occlusion mask \\
        \bottomrule[1.0pt]
     \end{tabular}
   \end{center}
\label{tab: notation}
\end{table}

\begin{figure*}[t]
  \centering
  \includegraphics[width=.98\textwidth]{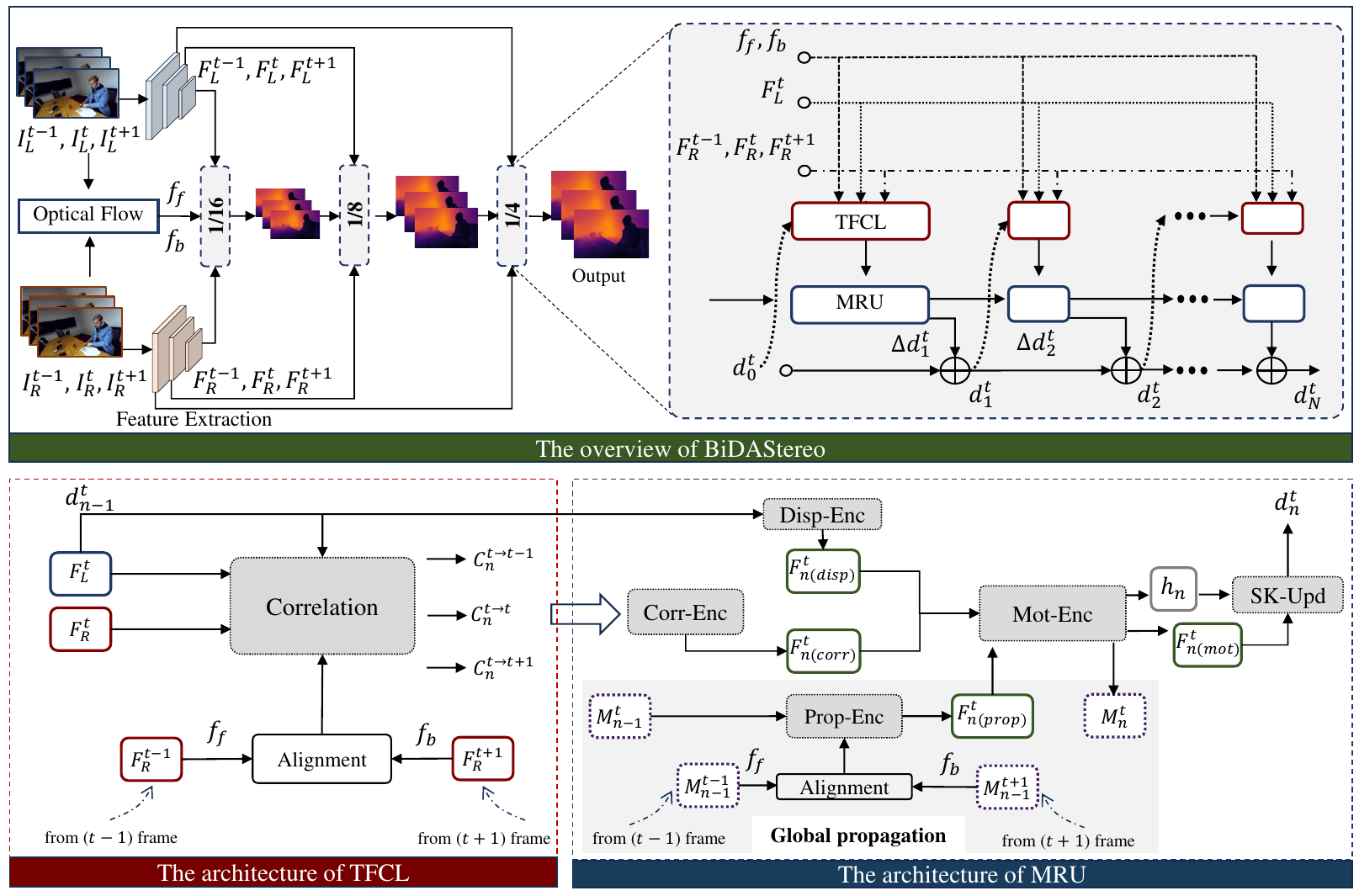}
  \vspace{-.5em}
  \caption{\textbf{Upper Left:} The overall pipeline of the proposed BiDAStereo. Given a pair of stereo sequences, bidirectional optical flows are estimated, and feature maps are extracted at three scales. At each scale, the predicted disparities are refined iteratively in the update module, and the final output of the previous stage is used as initialization for the next stage. The same update module is reused at each stage. \textbf{Upper Right:} The architecture of the update module. For each iteration, the Triple-Frame Correlation Layer (TFCL) computes cost volumes from triple-frame feature maps. The Motion-Propagation Recurrent Unit (MRU) is used for global cost aggregation and disparity estimation.  \textbf{Lower:} The architectures of TFCL and MRU. For TFCL, bidirectional alignment is conducted from the adjacent right frames to the center frame, and cost volumes are built between the left frame and the aligned right frame. For MRU, convolutional encoders are used for correlations, disparities, and motion features. A motion hidden state feature is introduced for each frame as the auxiliary context in global propagation. In each iteration, adjacent motion hidden state features are aligned towards the center one, updating the center feature and propagating wider temporal information.}
  \label{fig:framework}
\end{figure*}

\subsection{South Kensington Stereo Video Dataset} 
\textcolor{black}{As discussed in Sec.~\ref{sec:introduction}, finding a comprehensive dataset of real-world stereo videos that cover diverse scenarios remains challenging. One alternative is sourcing videos labeled as ``stereo" from platforms like YouTube, though these are often not well rectified. For example, in the construction of the VDW dataset \cite{Wang_2023_ICCV}, most videos were collected from YouTube and filtered using optical flow calculations, excluding those where over 10\% of pixels exhibited vertical disparity greater than two pixels. However, this approach does not guarantee proper rectification.}

\textcolor{black}{To address the lack of well-rectified data and offer a more reliable dataset for visualization and qualitative evaluation, we present the \textit{South Kensington SV} dataset, a real-world stereo videos covering indoor and outdoor scenes recorded via a ZED 2 stereo camera. %Handheld recording typically introduces unwanted shaking and jitter from body movement during walking. To counter this, we utilized an Insta360 Flow gimbal to stabilize the recordings and designed a custom 3D-printed camera holder to securely attach the camera to the gimbal. 
Videos were captured across locations in South Kensington, London, including the Imperial College campus, Hyde Park, and Chelsea. The dataset includes diverse urban scenarios with dynamic objects, encompassing various weather conditions (sunny, cloudy, rainy, night, etc) and camera movements, primarily from walking.}

\textcolor{black}{Fig.~\ref{fig:southken} shows some examples from the dataset. It consists of 266 stereo videos, each between 10 and 70 seconds in length, recorded at 1280x720 resolution and 30 fps. The raw videos were inspected to select sequences with consistent exposure without abrupt changes in recording parameters. The data collection process spanned over three months, amounting to approximately 300 man-hours. To validate the diversity of scenes and objects, we applied semantic segmentation using Mask2Former \cite{cheng2022masked} and object detection using YOLOv8 \cite{reis2023real}. Fig.~\ref{fig:bubble} presents a word cloud of the categories identified. In addition to its value for stereo video tasks, this dataset is also suitable for monocular video depth with the pseudo labels provided, and low-level vision tasks such as stereo video super-resolution, denoising, and compression.}

\section{Methodology}

Given a pair of rectified stereo sequences $\{\mathbf{I}^{t}_{L}, \mathbf{I}^{t}_{R}\}_{t\in (1,T)} \in \mathbb{R}^{H\times W\times 3} $ as input, the task of video stereo matching is to estimate a sequence of disparity maps $\{\mathbf{d}^{t}\}_{t\in (1,T)} \in \mathbb{R}^{H\times W} $ aligned with the left one, where $T$ is the number of frames. The challenge lies in devising a method capable of maintaining temporal consistency across the entire sequence. In this section, we illustrate how to achieve temporal consistency based on bidirectional alignment. We introduce a video-based network, BiDAStereo, which directly outputs a disparity sequence. For existing trained image-based methods, we then propose a plugin stabilizer network, BiDAStabilizer, to enhance their temporal consistency.

\subsection{BiDAStereo} \label{sec:BiDAStereo}
\textcolor{black}{Existing video-based methods often treat long sequences as separate clips, applying window-based operations across arbitrarily set number of frames~\cite{karaev2023dynamicstereo}. This approach fails to leverage the full length of the sequence. Additionally, these methods neglect the crucial alignment operation between adjacent frames. Given that matching points in stereo images shift over time, relying solely on temporal attention without alignment is sub-optimal. To address this limitation, we introduce bidirectional alignment in correlation and aggregation for video stereo matching. As illustrated in Fig.~\ref{fig:framework}, our framework comprises three modules: a feature extraction module, an optical flow module, and an update module for disparity estimations.}

\textcolor{black}{To simplify the presentation, we exemplify the input with three frames (the center and its neighboring frames) but the same approach can be applied to more than three frames. In the upper left part of Fig.~\ref{fig:framework}, input stereo sequences are processed by two shared-weight convolutional feature extraction modules. Multi-scale feature maps $\left \{\mathbf{F}_{L}^t ,\mathbf{F}_{R}^t \right \}_s \in R^{sH\times sW\times C}$ are extracted, where $s \in \left \{1/16, 1/8, 1/4\right \}$ represent down-sampled scales, and $C$ is the number of channels. Optical flow is estimated and downsampled to the corresponding resolution of the feature maps. Subsequently, the extracted features and the optical flow maps traverse three cascaded stages of the update module, which consists of a Triple-Frame Correlation Layer (TFCL) and a Motion-propagation Recurrent Unit (MRU). In TFCL, cost volumes are built using the adjacent aligned features and input into the MRU, iteratively refining disparity predictions. Initialized from a blank disparity map, the output disparities from the previous update stage are fed into the next stage. The same update module is used in each stage to reduce overall parameters. Finally, the predicted disparities at the last stage are upscaled to the original resolution.}

\subsubsection{Triple-Frame Correlation Layer} \label{Triple-Frame Correlation Layer}

\textcolor{black}{\textbf{Bidirectional alignment.} \textcolor{black}{Building cost volumes across multiple frames can enrich scene perception, since occluded points in one frame may be visible in adjacent frames. Directly extending per-frame correlation to a temporal version, for example, by establishing multi-frame correlation, is a naive and sub-optimal approach to cost aggregation, as unaligned features from different frames introduce noise. Alignment is therefore indispensable to ensure that corresponding matching points across frames are spatially consistent.} As shown in the lower left part of Fig.~\ref{fig:framework}, given the bidirectional optical flow maps in corresponding resolution $\{\mathbf{f}_f, \mathbf{f}_b\}$, the right features $\left \{\mathbf{F}_{R}^{t-1}, \mathbf{F}_{R}^{t+1} \right \}$ are first aligned towards the center frame $\left \{\mathbf{F}_{R}^{t}\right \}$ via alignment operation $\mathcal{A}(\cdot)$ to obtain aligned features $\left \{\hat{\mathbf{F}}_{R}^{t-1}, \hat{\mathbf{F}}_{R}^{t+1} \right \}$ with $\mathcal{A}(\cdot)$  implemented as bilinear grid sampling,
\begin{equation*}
   \left\{
    \begin{aligned}
    & \hat{\mathbf{F}}_{R}^{t-1} = \mathcal{A}(\mathbf{F}_{R}^{t-1}, \mathbf{f}_f), \\
    & \hat{\mathbf{F}}_{R}^{t+1} = \mathcal{A}(\mathbf{F}_{R}^{t+1}, \mathbf{f}_b).\\
    \end{aligned}
    \right.
\end{equation*}
}

\textcolor{black}{\textbf{Correlation.} Following bidirectional alignment, cost volumes are constructed employing the local correlation mechanism \cite{li2022practical}. We extend this mechanism to a triple-frame version. In particular, for the $n$-th iteration, the intermediate disparity map $\mathbf{d}^t_{n-1}$ from the previous iteration is employed to align right features towards the left feature map: 
\begin{equation*}
    \left\{
    \begin{aligned}
    \tilde{\mathbf{F}}_{R}^{t-1} &= \mathcal{A}(\hat{\mathbf{F}}_{R}^{t-1}, \mathbf{d}^t_{n-1}), \\
    \tilde{\mathbf{F}}_{R}^{t} &= \mathcal{A}({\mathbf{F}}_{R}^{t}, \mathbf{d}^t_{n-1}), \\
    \tilde{\mathbf{F}}_{R}^{t+1} &= \mathcal{A}(\hat{\mathbf{F}}_{R}^{t+1}, \mathbf{d}^t_{n-1}).\\
    \end{aligned}
    \right.
\end{equation*}
Then, the cost volumes $\{\mathbf{C}_{n}^{t \to t-1}(\bm{p}), \mathbf{C}_{n}^{t \to t}(\bm{p}), \mathbf{C}_{n}^{t \to t+1}(\bm{p})\}$ between the left feature  map $\mathbf{F}_{L}^{t}$ and the aligned right features $\left \{\tilde{\mathbf{F}}_{R}^{t-1}, \tilde{\mathbf{F}}_{R}^{t}, \tilde{\mathbf{F}}_{R}^{t+1} \right \}$ at position $\mathbf{p}$ can be formulated as follows,
\begin{equation*}
    \left\{
    \begin{aligned}
    \mathbf{C}_{n}^{t \to t-1}(\bm{p}) &= \underset{\mathbf{r} \in \mathbf{R}}{\text{Concat}} \{ \langle \mathbf{F}_{L}^{t}(\mathbf{p}) \cdot  \tilde{\mathbf{F}}_{R}^{t-1}(\mathbf{p} + \mathbf{r}) \rangle \}, \\
    \mathbf{C}_{n}^{t \to t}(\bm{p}) &= \underset{\mathbf{r} \in \mathbf{R}}{\text{Concat}} \{ \langle \mathbf{F}_{L}^{t}(\mathbf{p}) \cdot  \tilde{\mathbf{F}}_{R}^{t}(\mathbf{p} + \mathbf{r}) \rangle \}, \\
    \mathbf{C}_{n}^{t \to t+1}(\bm{p}) &= \underset{\mathbf{r} \in \mathbf{R}}{\text{Concat}} \{ \langle \mathbf{F}_{L}^{t}(\mathbf{p}) \cdot  \tilde{\mathbf{F}}_{R}^{t+1}(\mathbf{p} + \mathbf{r}) \rangle \}.\\
    \end{aligned}
    \right.
\end{equation*}
$\mathbf{R}$ is the search range of the current pixel, where $(\pm 4,0)$ and $(\pm 1,\pm 1)$ are alternatively adopted for horizontal and vertical directions in practical implementation. $\left \langle \cdot \right \rangle $ represents channel-wise product operation. The added degree of freedom in the vertical search range, similar to CREStereo \cite{li2022practical}, is for imperfect rectification in stereo pairs.}

\textcolor{black}{It is worth noting that the adjacent left frames are not used and only the center left frame is utilized in the correlation layer. In multiple-frame matching, constructing cost volumes across different source (right) frames increases potential matching options as auxiliary information. Conversely, considering multiple reference (left) frames merely results in ambiguity, lacking a clear physical location of features and potentially introducing noise that can impact the matching performance. This is further corroborated by the ablation study in Section \ref{Ablation Studies}.}

\subsubsection{Motion-propagation Recurrent Unit} \label{Motion-propagation based Recurrent Unit}

We propose a motion-propagation mechanism designed to exploit contextual information from the entire sequence, illustrated in Fig.~\ref{fig:framework}. During the update iteration, a motion hidden state $\mathbf{M}$ is introduced to cache the state for each frame. \textcolor{black}{This involves bidirectional alignment of neighboring states to iteratively update each frame. Unlike methods based on straightforward temporal extensions with 3D convolutions, information in BiDAStereo is not restricted to a local sliding window but instead undergoes recurrent propagation across the entire sequence.} Detailed elaboration on each module will be provided in subsequent subsections.

\textcolor{black}{\textbf{Motion propagation.} A motion hidden state feature $\mathbf{M}^{t} \in R^{sH\times sW\times C_0}$ is introduced for the $t$-th frame in the MRU, where $C_0$ is the channel number. For the first iteration at the first stage, $\mathbf{M}^{t}_{0}$ is randomly initialized. For the $n$-th iteration, the motion hidden state features $\mathbf{M}_{n-1}^{t-1}$ and $\mathbf{M}_{n-1}^{t+1}$ from adjacent frames are aligned towards the center one $\mathbf{M}_{n-1}^{t}$ via bidirectional optical flow maps $\{\mathbf{f}_f, \mathbf{f}_b\}$, formulated as follows, 
\begin{equation*}
   \left\{
    \begin{aligned}
    & \hat{\mathbf{M}}_{n-1}^{t-1} = \mathcal{A}(\mathbf{M}_{n-1}^{t-1}, \mathbf{f}_f),\\
    & \hat{\mathbf{M}}_{n-1}^{t+1} = \mathcal{A}(\mathbf{M}_{n-1}^{t+1}, \mathbf{f}_b).\\
    \end{aligned}
    \right.
\end{equation*}
Then, the aligned features $\{\hat{\mathbf{M}}_{n-1}^{t-1}, \hat{\mathbf{M}}_{n-1}^{t+1}\}$  and the center one $\mathbf{M}_{n-1}^{t}$ are fused by a motion-propagation encoder to obtain the motion propagation feature $\mathbf{F}_{n (prop)}^{t}$:
\begin{equation*}
\mathbf{F}_{n (prop)}^{t} = \text{Prop-Enc}(\hat{\mathbf{M}}_{n-1}^{t-1}, \mathbf{M}_{n-1}^{t}, \hat{\mathbf{M}}_{n-1}^{t+1}).
\end{equation*}
Across various update scales, the motion hidden state is not re-initialized, and the final output hidden state from the previous level is up-sampled as the initial input for the next level.}

\textcolor{black}{\textbf{Encoder blocks.} After building the cost volumes, the correlation feature $\mathbf{F}_{n (corr)}^{t}$ is obtained using a convolutional correlation encoder (Corr-Enc):
\begin{equation*}
\begin{aligned}
\mathbf{F}_{n (corr)}^{t} &= {\text{Corr-Enc}}(\mathbf{C}_{n}^{t \to t-1}, \mathbf{C}_{n}^{t \to t}, \mathbf{C}_{n}^{t \to t+1}).
\end{aligned}
\end{equation*}
The disparity encoder (Disp-Enc) is applied to get $\mathbf{F}_{n (disp)}^{t}$:
\begin{equation*}
\begin{aligned}
\mathbf{F}_{n (disp)}^{t} &= \text{Disp-Enc}(\mathbf{d}_{n-1}^{t}).
\end{aligned}
\end{equation*}
Subsequently, these two features are concatenated with the motion propagation feature $\mathbf{F}_{n (prop)}^{t}$ and processed by the motion encoder (Mot-Enc) to generate the motion feature $\mathbf{F}_{n (mot)}^{t}$, hidden state $\mathbf{h}_n$ \cite{lipson2021raft}, and the updated motion hidden state feature $\mathbf{M}_{n}^{t}$:
\begin{equation*}
\begin{aligned}
\mathbf{F}_{n (mot)}^{t}, \mathbf{h}_n, \mathbf{M}_{n}^{t} &= \text{Mot-Enc}(\mathbf{F}_{n (corr)}^{t}, \mathbf{F}_{n (disp)}^{t}, \mathbf{F}_{n (prop)}^{t}).
\end{aligned}
\end{equation*} }

\textcolor{black}{\textbf{Super kernel updater.} The hidden state $\mathbf{h}_n$ and the motion feature $\mathbf{F}_{n (mot)}^{t}$ are passed through a super kernel updater to get the predicted disparities residuals $\Delta \mathbf{d}_n^t$, which are added to the current disparities $\mathbf{d}_{n-1}^t$ and obtained $\mathbf{d}_{n}^t$, iteratively refining the disparity predictions. Following \cite{karaev2023dynamicstereo}, 3D convolutions are used in the updater for video sequences to enhance temporal consistency. Different from the regular updater with kernel size $1 \times 1 \times 5$, we introduce an extra convolution layer with a super kernel size $1 \times 1 \times 15$. The proposed layer is only adopted for the horizontal direction to expand the receptive field along the epipolar line.}

\begin{figure*}[tbp]
   \begin{center}
   \includegraphics[width=.98\linewidth]{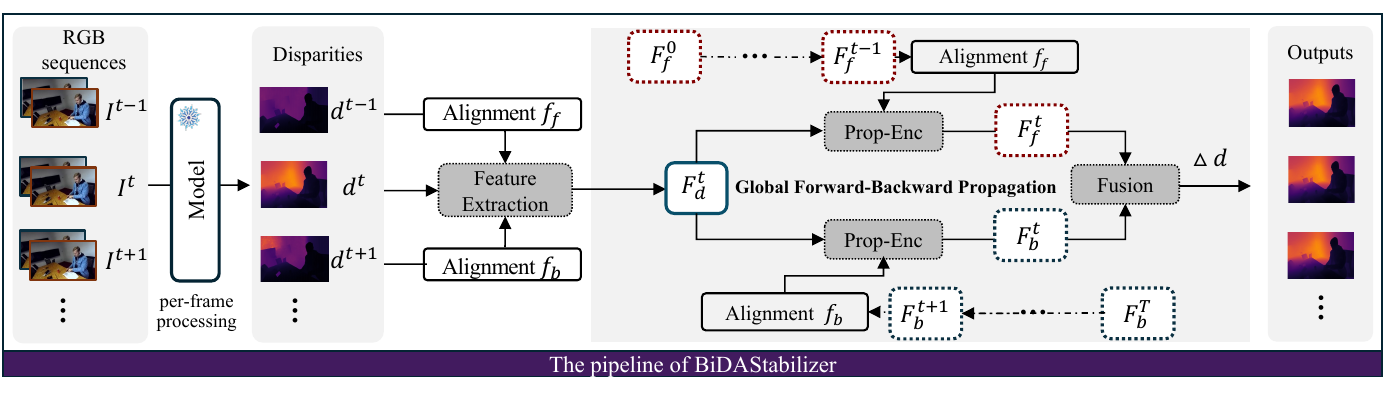}
   \end{center}
    \vspace{-.5em}
      \caption{Overview of the BiDAStabilizer pipeline. Given a pair of stereo sequences, disparities are estimated using a frozen image-based model. Bidirectional optical flows are estimated and adjacent disparities are aligned accordingly. Forward ${F}_f$ and backward  ${F}_b$ hidden state features are integrated with disparity features $F_d$ via propagation encoder for global consistency. For each central frame, the adjacent hidden state features are aligned with it, and then used to update the central feature and propagate temporal information throughout the sequence.} 
   \label{fig:bidastabilizer}
   \vspace{0em}
\end{figure*}

\subsection{BiDAStabilizer} \label{sec:BiDAStabilizer}

Most existing stereo methods focus on matching image pairs rather than processing videos. Training a stereo model for videos from scratch is both time- and resource-intensive. Given the availability of numerous pre-trained models in the model zoo, we propose a plugin stabilizer network to utilize these weights and extend the applicability of image-based methods to temporally consistent video stereo matching. 

As shown in Fig.~\ref{fig:bidastabilizer}, given a stereo sequence $\{\mathbf{I}^{t}_{L}, \mathbf{I}^{t}_{R}\}_{t\in (1,T)} $ as input, we first obtain a sequence of disparity maps $\{\mathbf{d}^{t}\}_{t\in (1,T)} $ frame by frame using a frozen image-based model. Simultaneously, bidirectional optical flow $\{\mathbf{f}_f, \mathbf{f}_b\} \in \mathbb{R}^{H\times W\times 2}$ are estimated from the input sequence. Since the RGB and disparity spaces are aligned, the adjacent disparities $\{\mathbf{d}^{t-1}, \mathbf{d}^{t+1}\}$ can be aligned with the central frame disparity via  alignment operation $\mathcal{A}(\cdot)$ using the optical flows $\{\mathbf{f}_f, \mathbf{f}_b\}$, as formulated below:
\begin{equation}
   \left\{
    \begin{aligned}
    & \hat{\mathbf{d}}^{t-1} = \mathcal{A}(\mathbf{d}^{t-1}, \mathbf{f}_f), \\
    & \hat{\mathbf{d}}^{t+1} = \mathcal{A}(\mathbf{d}^{t+1}, \mathbf{f}_b).\\
    \end{aligned}
    \right.
\end{equation}
These aligned disparities and the central disparity are then processed through a convolutional feature extraction module (FE) to produce the disparity feature map $\mathbf{F}_d^t$:
\begin{equation}
     \mathbf{F}_d^t = \text{FE}(\hat{\mathbf{d}}^{t-1}, \mathbf{d}^{t}, \hat{\mathbf{d}}^{t+1}).
\end{equation}
With the adjacent alignment, the stabilizer network can utilize temporal information from neighboring two frames. However, the extent of information propagation is still confined to a local range. To facilitate the propagation of global consistency, it's essential to establish a state variable representing each frame and fuse it across adjacent frames. While feature maps from images are commonly used as state variables in some low-level tasks, they are not well-suited for stereo matching, as fusion with other frames could significantly impact matching accuracy. Instead, separate hidden state features, which are more appropriate for global propagation and dynamic scenes, are introduced to maintain the state of each frame and update it in subsequent frames. 

For the forward process, the forward hidden state feature $\mathbf{F}^{0}_f$ for the first frame is initialized to zero. For the $t$-th frame, the forward hidden state feature $\mathbf{F}_{f}^{t-1}$ from the previous frame is aligned to the central frame and then combined with the disparity feature $\mathbf{F}_d^t$, which is processed by the propagation encoder (Prop-Enc) to obtain the updated hidden state feature $\mathbf{F}_{f}^{t}$:

% As described in Sec.~\ref{Motion-propagation based Recurrent Unit}, to facilitate global consistency propagation, hidden state features are introduced to maintain the state of each frame and update it in subsequent frames. For the forward process, the forward hidden state feature $\mathbf{F}^{0}_f$ for the first frame is initialized to zero. For the $t$-th frame, the forward hidden state feature $\mathbf{F}_{f}^{t-1}$ from the previous frame is aligned to the central frame and then combined with the disparity feature $\mathbf{F}_d^t$, which is processed by the propagation encoder (Prop-Enc) to obtain the updated hidden state feature $\mathbf{F}_{f}^{t}$:

\begin{equation}
    \left\{
    \begin{aligned}
    & \hat{\mathbf{F}}_{f}^{t-1} = \mathcal{A}(\mathbf{F}_{f}^{t-1}, \mathbf{f}_f), \\
    & \mathbf{F}_{f}^{t} = \text{Prop-Enc}(\hat{\mathbf{F}}_{f}^{t-1}, \mathbf{F}_d^t).\\
    \end{aligned}
    \right.
\end{equation}
Similarly, the same operation is adopted for the backward process from frame $T$ to $0$, where $\mathbf{F}^{T}_b$ is initialized with zero:
\begin{equation}
    \left\{
    \begin{aligned}
    & \hat{\mathbf{F}}_{b}^{t+1} = \mathcal{A}(\mathbf{F}_{b}^{t+1}, \mathbf{f}_b), \\
    & \mathbf{F}_{b}^{t} = \text{Prop-Enc}(\hat{\mathbf{F}}_{b}^{t+1}, \mathbf{F}_d^t).\\
    \end{aligned}
    \right.
\end{equation}
Finally, the forward and backward propagated feature maps $\{ \mathbf{F}_{f}^{t}, \mathbf{F}_{b}^{t} \}$ are fused using a convolutional fusion module to obtain the output residual $\Delta \mathbf{d}^{t}$:
\begin{equation}
     \Delta \mathbf{d}^{t} = \text{Fusion}(\mathbf{F}_{f}^{t}, \mathbf{F}_{b}^{t}),
\end{equation}
which is then added to the original input disparity maps.

\begin{table*}[tb] 
\caption{In-domain evaluation on indoor Dynamic Replica (DR) \cite{karaev2023dynamicstereo}, outdoor KITTI Depth (K) \cite{kittidepth}, and the proposed outdoor Infinigen Stereo Video (ISV). SF - SceneFlow \cite{mayer2016large}.  The percentage change is shown in color, with green indicating improvement.}
% \vspace{-.5em}
\centering
\setlength{\tabcolsep}{6.pt}
\footnotesize
\begin{tabular}{c|c|cccccccccc}
\toprule[1.0pt]
\rowcolor{mygray}
Training &    & \multicolumn{10}{c}{Dynamic Replica}  \\
\rowcolor{mygray}
Data & \multirow{-2}{*}{Method} & OPW\(_{100}\)$\downarrow $ & OPW\(_{30}\)$\downarrow $ & RTC$\uparrow $ & TCC$\uparrow $ & TCM$\uparrow $ & $\delta_{1px}^{t}$$\downarrow $ & $\delta_{3px}^{t}$$\downarrow $ & TEPE$\downarrow $ & EPE$\downarrow $ & $\delta_{1px}$$\downarrow $ \\
\midrule
\multirow{9}{*}{SF + DR}        & RAFTStereo \cite{lipson2021raft}     & 0.014	& 0.014 & 0.930 & 0.771 & 0.416 & 0.84  & 0.27 & 0.082 & 0.15 & 1.88 \\
                                & R. + BiDAStabilizer (ours)   & 0.007 & 0.007 & 0.980 & 0.894 & 0.487 & 0.60 & 0.22 & 0.046 & 0.14 & 1.78 \\
                                &    & \scriptsize \textcolor{d-green}{-50.0\%} & \scriptsize \textcolor{d-green}{-50.0\%} & \scriptsize \textcolor{d-green}{+5.38\%} & \scriptsize \textcolor{d-green}{+16.0\%} & \scriptsize \textcolor{d-green}{+17.1\%} & \scriptsize \textcolor{d-green}{-28.6\%} & \scriptsize \textcolor{d-green}{-18.5\%} & \scriptsize \textcolor{d-green}{-43.9\%} & \scriptsize \textcolor{d-green}{-6.67\%} & \scriptsize \textcolor{d-green}{-5.32\%} \\ 
                                
\cmidrule(lr){2-12}
                                & IGEVStereo \cite{xu2023iterative}   & 0.012 & 0.012 & 0.935 & 0.788 & 0.447 & 0.77 & 0.23 & 0.076 & 0.15 & 1.91 \\
                                & {I. + BiDAStabilizer (ours)} & 0.006 & 0.006 & 0.982 & 0.900 & 0.508 & 0.54 & 0.19 & 0.043 & 0.14 & 1.83 \\
                                &    & \scriptsize \textcolor{d-green}{-50.0\%} & \scriptsize \textcolor{d-green}{-50.0\%} & \scriptsize \textcolor{d-green}{+5.03\%} & \scriptsize \textcolor{d-green}{+14.2\%} & \scriptsize \textcolor{d-green}{+13.6\%} & \scriptsize \textcolor{d-green}{-29.9\%} & \scriptsize \textcolor{d-green}{-17.4\%} & \scriptsize \textcolor{d-green}{-43.4\%} & \scriptsize \textcolor{d-green}{-6.67\%} & \scriptsize \textcolor{d-green}{-4.19\%} \\
\cmidrule(lr){2-12}
                                & DynamicStereo \cite{karaev2023dynamicstereo} & 0.013	& 0.013 & 0.948 & 0.770 & 0.397 & {0.68}    & {0.23} & {0.075} & 0.21 & 3.32   \\
                                & {BiDAStereo (ours)}  & 0.010 & 0.010 & 0.970 & 0.804 & 0.457 & {0.61}    & {0.22}  & {0.062} & 0.22 & {2.81}   \\
                                &    & \scriptsize \textcolor{d-green}{-23.1\%} & \scriptsize \textcolor{d-green}{-23.1\%} & \scriptsize \textcolor{d-green}{+2.32\%} & \scriptsize \textcolor{d-green}{+4.42\%} & \scriptsize \textcolor{d-green}{+15.1\%} & \scriptsize \textcolor{d-green}{-10.3\%} & \scriptsize \textcolor{d-green}{-4.35\%} & \scriptsize \textcolor{d-green}{-17.3\%} & \scriptsize \textcolor{d-red}{+4.76\%} & \scriptsize \textcolor{d-green}{-15.4\%} \\ 
\midrule
\midrule
\rowcolor{mygray}
Training &    & \multicolumn{10}{c}{KITTI Depth}  \\
\rowcolor{mygray}
Data & \multirow{-2}{*}{Method} & OPW\(_{100}\)$\downarrow $  & OPW\(_{30}\)$\downarrow $ & RTC$\uparrow $ & TCC$\uparrow $ & TCM$\uparrow $ & $\delta_{1px}^{t}$$\downarrow $ & $\delta_{3px}^{t}$$\downarrow $ & TEPE$\downarrow $ & EPE$\downarrow $ & $\delta_{1px}$$\downarrow $ \\
\midrule
\multirow{9}{*}{SF + K}        & RAFTStereo \cite{lipson2021raft}     & 6.472 & 5.524 & 0.034 & 0.878 & 0.976 & 6.53 &0.18 & 0.36 & 0.41 & 6.13	\\
                                & {R. + BiDAStabilizer (ours)}       & 6.520 & 5.550 & 0.033 & 0.881 & 0.976 & 6.29 & 0.18 & 0.34 &  0.41 & 5.72 \\
                                &    & \scriptsize \textcolor{d-red}{+0.74\%} & \scriptsize \textcolor{d-red}{+0.47\%} & \scriptsize \textcolor{d-red}{-2.94\%} & \scriptsize \textcolor{d-green}{+0.34\%} & \scriptsize \textcolor{gray}{0.00\%} & \scriptsize \textcolor{d-green}{-3.68\%} & \scriptsize \textcolor{gray}{0.00\%} & \scriptsize \textcolor{d-green}{-5.56\%} & \scriptsize \textcolor{gray}{0.00\%} & \scriptsize \textcolor{d-green}{-6.69\%} \\ 
\cmidrule(lr){2-12}
                                & IGEVStereo \cite{xu2023iterative}      & 6.523 & 5.549 & 0.033 & 0.874 & 0.975 & 7.11 & 0.29 & 0.38 & 0.44 & 6.87 \\
                                & {I. + BiDAStabilizer (ours)}       & 6.505 & 5.542 & 0.033 & 0.879 & 0.976 & 6.61 & 0.28 & 0.35 & 0.43 & 6.64 \\
                                &    & \scriptsize \textcolor{d-green}{-0.28\%} & \scriptsize \textcolor{d-green}{-0.13\%} & \scriptsize \textcolor{gray}{0.00\%} & \scriptsize \textcolor{d-green}{+0.57\%} & \scriptsize \textcolor{d-green}{+0.10\%} & \scriptsize \textcolor{d-green}{-7.03\%} & \scriptsize \textcolor{d-green}{-3.45\%} & \scriptsize \textcolor{d-green}{-7.89\%} & \scriptsize \textcolor{d-green}{-2.27\%} & \scriptsize \textcolor{d-green}{-3.35\%} \\ 
\cmidrule(lr){2-12}
                                & DynamicStereo \cite{karaev2023dynamicstereo} & 6.455 & 5.521 & 0.034 & 0.878 & 0.976 & 6.64 & 0.20 & 0.36 & 0.42 & 6.45\\
                                & {BiDAStereo (ours)}            & 6.482 & 5.526 & 0.033 & 0.880 & 0.977 & 6.20 & 0.16 & 0.34 & 0.39 & 5.65 \\
                                &    & \scriptsize \textcolor{d-red}{+0.42\%} & \scriptsize \textcolor{d-red}{+0.09\%} & \scriptsize \textcolor{d-red}{-2.94\%} & \scriptsize \textcolor{d-green}{+0.23\%} & \scriptsize \textcolor{d-green}{+0.10\%} & \scriptsize \textcolor{d-green}{-6.63\%} & \scriptsize \textcolor{d-green}{-20.00\%} & \scriptsize \textcolor{d-green}{-5.56\%} & \scriptsize \textcolor{d-green}{-7.14\%} & \scriptsize \textcolor{d-green}{-12.40\%} \\ 
                                
\midrule
\midrule
\rowcolor{mygray}
Training &    & \multicolumn{10}{c}{Infinigen Stereo Video}  \\
\rowcolor{mygray}
Data & \multirow{-2}{*}{Method} & OPW\(_{100}\)$\downarrow $  & OPW\(_{50}\)$\downarrow $ & RTC$\uparrow $ & TCC$\uparrow $ & TCM$\uparrow $ & $\delta_{1px}^{t}$$\downarrow $ & $\delta_{3px}^{t}$$\downarrow $ & TEPE$\downarrow $ & EPE$\downarrow $ & $\delta_{3px}$$\downarrow $ \\
\midrule
\multirow{9}{*}{SF + ISV}  & RAFTStereo \cite{lipson2021raft}     & 1.439 & 1.184 & 0.646 & 0.262 & 0.246 & 17.04 & 9.12 & 2.04 & 2.73 & 11.52 \\
                                & {R. + BiDAStabilizer (ours)}       & 1.056 & 0.810 & 0.802 & 0.325 & 0.271 & 14.43 & 8.00 & 1.74 & 2.65 & 11.44 \\
                                &    & \scriptsize \textcolor{d-green}{-26.6\%} & \scriptsize \textcolor{d-green}{-31.6\%} & \scriptsize \textcolor{d-green}{+24.1\%} & \scriptsize \textcolor{d-green}{+24.0\%} & \scriptsize \textcolor{d-green}{+10.2\%} & \scriptsize \textcolor{d-green}{-15.3\%} & \scriptsize \textcolor{d-green}{-12.3\%} & \scriptsize \textcolor{d-green}{-14.7\%} & \scriptsize \textcolor{d-green}{-2.93\%} & \scriptsize \textcolor{d-green}{-0.69\%} \\ 
\cmidrule(lr){2-12}
                                & IGEVStereo \cite{xu2023iterative}      & 5.259 & 4.948 & 0.641 & 0.277 & 0.237 & 16.78 & 8.93 & 1.83 & 2.32 & 10.56 \\
                                & {I. + BiDAStabilizer (ours)} & 1.601 & 1.199 & 0.791 & 0.350 & 0.276 & 14.32 & 7.88 & 1.56 & 2.23 & 10.39 \\
                                &    & \scriptsize \textcolor{d-green}{-69.6\%} & \scriptsize \textcolor{d-green}{-75.8\%} & \scriptsize \textcolor{d-green}{+23.4\%} & \scriptsize \textcolor{d-green}{+26.4\%} & \scriptsize \textcolor{d-green}{+16.5\%} & \scriptsize \textcolor{d-green}{-14.7\%} & \scriptsize \textcolor{d-green}{-11.8\%} & \scriptsize \textcolor{d-green}{-14.8\%} & \scriptsize \textcolor{d-green}{-3.88\%} & \scriptsize \textcolor{d-green}{-1.61\%} \\ 
\cmidrule(lr){2-12}
                                & DynamicStereo \cite{karaev2023dynamicstereo} & 88.800 & 130.290 & 0.687 & 0.265 & 0.241 & 15.61 & 8.44 & 1.80 & 3.98 & 12.17 \\
                                & {BiDAStereo (ours)}  & 0.912 & 0.565 & 0.699 & 0.270 & 0.251 & 15.76 & 8.33 & 1.81 & 3.40 & 14.63 \\
                                &    & \scriptsize \textcolor{d-green}{-99.0\%} & \scriptsize \textcolor{d-green}{-99.6\%} & \scriptsize \textcolor{d-green}{+1.75\%} & \scriptsize \textcolor{d-green}{+1.89\%} & \scriptsize \textcolor{d-green}{+4.15\%} & \scriptsize \textcolor{d-red}{+0.96\%} & \scriptsize \textcolor{d-green}{-1.30\%} & \scriptsize \textcolor{d-red}{+0.56\%} & \scriptsize \textcolor{d-green}{-14.6\%} & \scriptsize \textcolor{d-red}{+20.2\%} \\ 

\bottomrule[1.0pt]
\end{tabular}
\label{tab:in domain}
\end{table*}

\subsection{Loss Function} \label{sec:loss}
In the training process, $T$ frames are used, and for each frame, $N$ or $1$ disparity predictions are generated for  BiDAStereo or BiDAStabilizer, respectively. All disparity predictions are supervised in spatial accuracy by $l_1$ distance with ground truth disparities. The last disparity prediction is selected as the final output. The spatial loss $\mathcal{L}_s$ is formulated as follows:
\begin{equation}
\mathcal{L}_s = \sum_{t=1}^{T} \sum_{n=1}^{N} 
\gamma^{N - n} \left\| \mathbf{d}_{\mathrm{gt}}^{t} - \mathbf{d}_{n}^{t} \right\|,
\label{eq:spatial loss}
\end{equation}
where $\gamma = 0.9$ and $\mathbf{d}_{\mathrm{gt}}^{t}$ is the ground truth for the $t$-th frame. Upsampling is used for lower resolution disparities to the resolution of the ground truth. The temporal consistency loss $\mathcal{L}_c$ is formulated as:
\begin{equation}
    \mathcal{L}_c = \sum_{t=2}^{T-1} \{ \mathbf{O}_{t-1} \cdot \left\|  \hat{\mathbf{d}}^{t-1} -  \mathbf{d}^{t} \right\| + \mathbf{O}_{t+1} \cdot \left\|  \hat{\mathbf{d}}^{t+1} -  \mathbf{d}^{t} \right\| \},
\end{equation}
where $\mathbf{O}$ is the occlusion mask by optical flow, and $\{\hat{\mathbf{d}}^{t-1},  \hat{\mathbf{d}}^{t+1}\}$ are the aligned frames towards $\mathbf{d}^{t}$. The total loss is the weighted sum of these two losses:
\begin{equation}
    \mathcal{L}_{total} = \mathcal{L}_s  + \lambda \cdot \mathcal{L}_c,
\end{equation}
where $\lambda$ is set as 0 for BiDAStereo and 0.2 for BiDAStabilizer.

\section{Experiments}
% This section presents the evaluation datasets and implementation details, demonstrates its performance on temporal consistency, and performs ablation studies to confirm the effectiveness of its components.

\begin{table*}[htb] 
\caption{Out-of-domain evaluation on Sintel clean and final pass datasets \cite{sintel}. SF - SceneFlow \cite{mayer2016large}, DR - Dynamic Replica \cite{karaev2023dynamicstereo}.  The percentage change is shown in color beneath each result, with green indicating improvement.}
% \vspace{-.6em}
\centering
\setlength{\tabcolsep}{7.pt}
\footnotesize
\begin{tabular}{c|c|cccccccccc}
\toprule[1.0pt]
\rowcolor{mygray}
{Training}&    & \multicolumn{10}{c}{Sintel Clean}  \\
\rowcolor{mygray}
Data & \multirow{-2}{*}{Method} & OPW\(_{100}\)$\downarrow $  & OPW\(_{30}\)$\downarrow $ & RTC$\uparrow $ & TCC$\uparrow $ & TCM$\uparrow $ & $\delta_{1px}^{t}$$\downarrow $ & $\delta_{3px}^{t}$$\downarrow $ & TEPE$\downarrow $ & EPE$\downarrow $ & $\delta_{3px}$$\downarrow $ \\
\midrule
\multirow{9}{*}{SF}             & RAFTStereo \cite{lipson2021raft}  & 0.657 & 0.246 & 0.543 & 0.685 & 0.694 & 9.32  & 4.51 & 0.92 & 1.44 & 6.12  \\
                                &R. + BiDAStabilizer (ours)      & 0.609 & 0.232 & 0.583 & 0.704 & 0.706 & 9.22 & 4.42 & 0.86 & 1.44 & 6.29 \\
                                &    & \scriptsize \textcolor{d-green}{-7.31\%} & \scriptsize \textcolor{d-green}{-5.69\%} & \scriptsize \textcolor{d-green}{+7.37\%} & \scriptsize \textcolor{d-green}{+2.77\%} & \scriptsize \textcolor{d-green}{+1.73\%} & \scriptsize \textcolor{d-green}{-1.07\%} & \scriptsize \textcolor{d-green}{-2.00\%} & \scriptsize \textcolor{d-green}{-6.52\%} & \scriptsize \textcolor{gray}{0.00\%} & \scriptsize \textcolor{d-red}{+2.78\%} \\ 
\cmidrule(lr){2-12}
                                & IGEVStereo \cite{xu2023iterative}  & 0.699 & 0.257 & 0.547 & 0.693 & 0.718 & 8.84 & 4.61 & 1.13 & 1.43 & 6.02 \\
                                & {I. + BiDAStabilizer (ours)}     & 0.639 & 0.256 & 0.585 & 0.711 & 0.706 & 8.76 & 4.52 & 1.02 & 1.38 & 6.25  \\
                                &    & \scriptsize \textcolor{d-green}{-8.58\%} & \scriptsize \textcolor{d-green}{-0.39\%} & \scriptsize \textcolor{d-green}{+6.95\%} & \scriptsize \textcolor{d-green}{+2.60\%} & \scriptsize \textcolor{d-red}{-1.67\%} & \scriptsize \textcolor{d-green}{-0.90\%} & \scriptsize \textcolor{d-green}{-1.95\%} & \scriptsize \textcolor{d-green}{-9.73\%} & \scriptsize \textcolor{d-green}{-3.50\%} & \scriptsize \textcolor{d-red}{+3.82\%} \\ 
                                
\cmidrule(lr){2-12}
                                & DynamicStereo \cite{karaev2023dynamicstereo} & 0.473 & 0.181 & 0.558 & 0.684 & 0.679 & {8.41}  & {3.93} & {0.77} & 1.74 & {6.10} \\
                                & {BiDAStereo (ours)} & 0.481	& 0.216	& 0.564	& 0.688	& 0.709 & {8.29} & {3.79} & {0.73} & 1.26 &  {5.94} \\
                                &    & \scriptsize \textcolor{d-red}{+1.69\%} & \scriptsize \textcolor{d-red}{+19.3\%} & \scriptsize \textcolor{d-green}{+1.08\%} & \scriptsize \textcolor{d-green}{+0.58\%} & \scriptsize \textcolor{d-green}{+4.42\%} & \scriptsize \textcolor{d-green}{-1.43\%} & \scriptsize \textcolor{d-green}{-3.56\%} & \scriptsize \textcolor{d-green}{-5.19\%} & \scriptsize \textcolor{d-green}{-27.6\%} & \scriptsize \textcolor{d-green}{-2.62\%} \\ 
								
\midrule
\multirow{9}{*}{SF + DR}        & RAFTStereo \cite{lipson2021raft} & 0.631 & 0.232 & 0.544 & 0.690 & 0.708 & 9.07  & 4.40 & 0.89 & 1.20  & 5.83   \\
                                & {R. + BiDAStabilizer (ours)}     & 0.557 & 0.209 & 0.617 & 0.718 & 0.720 & 8.84 & 4.08 & 0.79  & 1.19 & 6.01  \\
                                &    & \scriptsize \textcolor{d-green}{-11.7\%} & \scriptsize \textcolor{d-green}{-9.91\%} & \scriptsize \textcolor{d-green}{+13.4\%} & \scriptsize \textcolor{d-green}{+4.06\%} & \scriptsize \textcolor{d-green}{+1.69\%} & \scriptsize \textcolor{d-green}{-2.54\%} & \scriptsize \textcolor{d-green}{-7.27\%} & \scriptsize \textcolor{d-green}{-11.2\%} & \scriptsize \textcolor{d-green}{-0.83\%} & \scriptsize \textcolor{d-red}{+3.09\%} \\ 
\cmidrule(lr){2-12}
                                & IGEVStereo \cite{xu2023iterative}      & 0.858 & 0.238 & 0.548 & 0.699 & 0.731 & 8.09 & 4.03 & 0.86 & 1.07 & 5.45 \\
                                & {I. + BiDAStabilizer (ours)} & 0.740 & 0.210 & 0.614 & 0.729 & 0.740 & 8.06 & 3.76 & 0.74 & 1.04 & 5.63 \\
                                &    & \scriptsize \textcolor{d-green}{-13.8\%} & \scriptsize \textcolor{d-green}{-11.8\%} & \scriptsize \textcolor{d-green}{+12.0\%} & \scriptsize \textcolor{d-green}{+4.29\%} & \scriptsize \textcolor{d-green}{+1.23\%} & \scriptsize \textcolor{d-green}{-0.37\%} & \scriptsize \textcolor{d-green}{-6.70\%} & \scriptsize \textcolor{d-green}{-14.0\%} & \scriptsize \textcolor{d-green}{-2.80\%} & \scriptsize \textcolor{d-red}{+3.30\%} \\ 
\cmidrule(lr){2-12}
                                & DynamicStereo \cite{karaev2023dynamicstereo} & 0.484 & 0.196 & 0.553 & 0.675 & 0.666 & {8.46}  & {3.93} & {0.76} & 1.37 & {5.77} \\
                                & {BiDAStereo (ours)} & 0.458	& 0.184	& 0.568	& 0.698	& 0.711 & {8.03}  & {3.76} & {0.75} & 1.30 & {5.75} \\
                                &    & \scriptsize \textcolor{d-green}{-5.37\%} & \scriptsize \textcolor{d-green}{-6.12\%} & \scriptsize \textcolor{d-green}{+2.71\%} & \scriptsize \textcolor{d-green}{+3.41\%} & \scriptsize \textcolor{d-green}{+6.76\%} & \scriptsize \textcolor{d-green}{-5.08\%} & \scriptsize \textcolor{d-green}{-4.33\%} & \scriptsize \textcolor{d-green}{-1.32\%} & \scriptsize \textcolor{d-green}{-5.11\%} & \scriptsize \textcolor{d-green}{-0.35\%} \\ 
                                
\midrule
\midrule
\rowcolor{mygray}
 {Training}&    & \multicolumn{10}{c}{Sintel Final}  \\
\rowcolor{mygray}
Data & \multirow{-2}{*}{Method} & OPW\(_{100}\)$\downarrow $  & OPW\(_{30}\)$\downarrow $ & RTC$\uparrow $ & TCC$\uparrow $ & TCM$\uparrow $ & $\delta_{1px}^{t}$$\downarrow $ & $\delta_{3px}^{t}$$\downarrow $ & TEPE$\downarrow $ & EPE$\downarrow $ & $\delta_{3px}$$\downarrow $ \\
\midrule
\multirow{9}{*}{SF}             & RAFTStereo \cite{lipson2021raft}  & 0.942	& 0.601 & 0.521 & 0.530 & 0.692 & 13.69 & 7.08 & 2.10 & 4.89 & 10.39 \\
                                & {R. + BiDAStabilizer (ours)}      & 0.834 & 0.541 & 0.563 & 0.664 & 0.700 & 13.00 & 6.67 & 1.93  &  4.86 & 10.48 \\
                                &    & \scriptsize \textcolor{d-green}{-11.5\%} & \scriptsize \textcolor{d-green}{-9.98\%} & \scriptsize \textcolor{d-green}{+8.06\%} & \scriptsize \textcolor{d-green}{+25.3\%} & \scriptsize \textcolor{d-green}{+1.16\%} & \scriptsize \textcolor{d-green}{-5.04\%} & \scriptsize \textcolor{d-green}{-5.79\%} & \scriptsize \textcolor{d-green}{-8.10\%} & \scriptsize \textcolor{d-green}{-0.61\%} & \scriptsize \textcolor{d-red}{+0.87\%} \\ 
\cmidrule(lr){2-12}
                                & IGEVStereo \cite{xu2023iterative} & 0.889	& 0.432	& 0.537	& 0.662	& 0.710 & 12.65	& 6.59 & 1.74 & 2.44 & 8.94 \\
                                & {I. + BiDAStabilizer (ours)}      & 0.789 & 0.387 & 0.575 & 0.681 & 0.712 & 11.84 & 6.16 & 1.57 & 2.41 & 9.12 \\
                                &    & \scriptsize \textcolor{d-green}{-11.2\%} & \scriptsize \textcolor{d-green}{-10.4\%} & \scriptsize \textcolor{d-green}{+7.07\%} & \scriptsize \textcolor{d-green}{+2.87\%} & \scriptsize \textcolor{d-green}{+0.28\%} & \scriptsize \textcolor{d-green}{-6.40\%} & \scriptsize \textcolor{d-green}{-6.53\%} & \scriptsize \textcolor{d-green}{-9.77\%} & \scriptsize \textcolor{d-green}{-1.23\%} & \scriptsize \textcolor{d-red}{+2.01\%} \\ 
\cmidrule(lr){2-12}
                                & DynamicStereo \cite{karaev2023dynamicstereo} & 0.879 & 0.333 & 0.551 & 0.655 & 0.685 & {11.80} & {5.85} & {1.43}  & 4.26 & {8.96}  \\
                                & {BiDAStereo (ours)} & 0.598 & 0.293 & 0.561 & 0.66 & 0.691 & {11.65} & {5.53} & {1.26} & 1.99 & {8.78} \\
                                &    & \scriptsize \textcolor{d-green}{-32.0\%} & \scriptsize \textcolor{d-green}{-12.0\%} & \scriptsize \textcolor{d-green}{+1.81\%} & \scriptsize \textcolor{d-green}{+0.76\%} & \scriptsize \textcolor{d-green}{+0.88\%} & \scriptsize \textcolor{d-green}{-1.27\%} & \scriptsize \textcolor{d-green}{-5.47\%} & \scriptsize \textcolor{d-green}{-11.9\%} & \scriptsize \textcolor{d-green}{-53.3\%} & \scriptsize \textcolor{d-green}{-2.01\%} \\ 
								
\midrule
\multirow{9}{*}{SF + DR}        & RAFTStereo \cite{lipson2021raft} & 0.777 & 0.461 & 0.523 & 0.648 & 0.697 & 13.56 & 7.02 & 1.91 & 4.52 & 9.83  \\
                                & {R. + BiDAStabilizer (ours)}      & 0.654 & 0.408 & 0.595 & 0.679 & 0.710 & 12.31 & 6.36 & 1.73 & 4.50 & 9.97 \\
                                &    & \scriptsize \textcolor{d-green}{-15.8\%} & \scriptsize \textcolor{d-green}{-11.5\%} & \scriptsize \textcolor{d-green}{+13.8\%} & \scriptsize \textcolor{d-green}{+4.78\%} & \scriptsize \textcolor{d-green}{+1.87\%} & \scriptsize \textcolor{d-green}{-9.22\%} & \scriptsize \textcolor{d-green}{-9.40\%} & \scriptsize \textcolor{d-green}{-9.42\%} & \scriptsize \textcolor{d-green}{-0.44\%} & \scriptsize \textcolor{d-green}{+1.42\%} \\ 
\cmidrule(lr){2-12}
                                & IGEVStereo \cite{xu2023iterative} & 0.800	& 0.364	& 0.535	& 0.666	& 0.707 & 12.06	& 6.09 & 1.50 & 2.11 & 8.24 \\
                                & {I. + BiDAStabilizer (ours)} & 0.669 & 0.320 & 0.601 & 0.696 & 0.715 & 11.19 & 5.57 & 1.35 & 2.11 & 8.40 \\
                                &    & \scriptsize \textcolor{d-green}{-16.4\%} & \scriptsize \textcolor{d-green}{-12.1\%} & \scriptsize \textcolor{d-green}{+12.3\%} & \scriptsize \textcolor{d-green}{+4.50\%} & \scriptsize \textcolor{d-green}{+1.13\%} & \scriptsize \textcolor{d-green}{-7.21\%} & \scriptsize \textcolor{d-green}{-8.54\%} & \scriptsize \textcolor{d-green}{-10.0\%} & \scriptsize \textcolor{gray}{0.00\%} & \scriptsize \textcolor{d-red}{+1.94\%} \\ 
\cmidrule(lr){2-12}
                                & DynamicStereo \cite{karaev2023dynamicstereo} & 0.55 & 0.243 & 0.544 & 0.647 & 0.666 & {11.93} & {5.92} & {1.43} & 3.19  & {8.69} \\
                                & {BiDAStereo (ours)} & 0.52 & 0.207 & 0.565 & 0.670 & 0.705 & {11.04} & {5.30} & {1.22} & 1.98 & {8.52}  \\
                                &    & \scriptsize \textcolor{d-green}{-5.45\%} & \scriptsize \textcolor{d-green}{-14.8\%} & \scriptsize \textcolor{d-green}{+3.86\%} & \scriptsize \textcolor{d-green}{+3.55\%} & \scriptsize \textcolor{d-green}{+5.86\%} & \scriptsize \textcolor{d-green}{-7.46\%} & \scriptsize \textcolor{d-green}{-10.5\%} & \scriptsize \textcolor{d-green}{-14.7\%} & \scriptsize \textcolor{d-green}{-37.9\%} & \scriptsize \textcolor{d-green}{-1.96\%} \\ 
                                
\bottomrule[1.0pt]
\end{tabular}
\label{tab:out of domain}
\end{table*}

\subsection{Datasets and Benchmarks}
\textcolor{black}{For training, we utilize  SceneFlow (SF) dataset \cite{mayer2016large},  Dynamic Replica (DR) training set \cite{karaev2023dynamicstereo},  KITTI Depth (K) dataset \cite{kittidepth}, and  Infinigen Stereo Video (ISV) training set. Our work focuses on videos with moving cameras and is not intended for single-frame inputs. Consequently, commonly used image benchmarks, such as Middlebury \cite{middlebury} and ETH3D \cite{eth3d}, are unsuitable.  KITTI 2012/2015 \cite{kitti} datasets contain only two frames per scene, not full videos, and DrivingStereo \cite{yang2019drivingstereo} consists of images, not videos. After contacting the authors, we confirmed that DrivingStereo is not intended for video analysis, and their raw video data is unavailable. Therefore, we follow the evaluation protocol of DynamicStereo \cite{karaev2023dynamicstereo}, including the Sintel clean and final pass datasets \cite{sintel}, and the Dynamic Replica test set. Additionally, we consider  KITTI Depth dataset and Infinigen Stereo Video test set. Due to memory limitations, we use the first 150 frames from each video in the Dynamic Replica test set and  Infinigen Stereo Video dataset, and the first 300 frames from  KITTI Depth dataset. For all these datasets except SceneFlow, we first obtain their depth maps and convert them to disparity using the camera pose.}

\begin{table*}[tb] 
\caption{Robustness evaluation on indoor Dynamic Replica (DR) \cite{karaev2023dynamicstereo}, outdoor KITTI Depth (K) \cite{kittidepth}, and the proposed outdoor Infinigen Stereo Video (ISV).  The percentage change is shown in color beneath each result, with green indicating improvement.}
% \vspace{-.6em}
\centering
\setlength{\tabcolsep}{7.pt}
\footnotesize
\begin{tabular}{c|c|cccccccccc}
\toprule[1.0pt]
\rowcolor{mygray}
Training &    & \multicolumn{10}{c}{Dynamic Replica}  \\
\rowcolor{mygray}
Data & \multirow{-2}{*}{Method} & OPW\(_{100}\)$\downarrow $ & OPW\(_{30}\)$\downarrow $ & RTC$\uparrow $ & TCC$\uparrow $ & TCM$\uparrow $ & $\delta_{1px}^{t}$$\downarrow $ & $\delta_{3px}^{t}$$\downarrow $ & TEPE$\downarrow $ & EPE$\downarrow $ & $\delta_{1px}$$\downarrow $ \\
\midrule
\multirow{9}{*}{\makecell{SF + DR \\ + K + ISV}}        & RAFTStereo \cite{lipson2021raft}     & 0.020 & 0.020 & 0.904 & 0.732 & 0.394 & 1.19 & 0.38 & 0.104 & 0.198 & 2.71\\
                                & {R. + BiDAStabilizer (ours)} & 0.008 & 0.008 & 0.979 & 0.878 & 0.485 & 0.65 & 0.22 & 0.051 & 0.187 & 2.58 \\
                                &    & \scriptsize \textcolor{d-green}{-60.0\%} & \scriptsize \textcolor{d-green}{-60.0\%} & \scriptsize \textcolor{d-green}{+8.30\%} & \scriptsize \textcolor{d-green}{+19.9\%} & \scriptsize \textcolor{d-green}{+23.1\%} & \scriptsize \textcolor{d-green}{-45.4\%} & \scriptsize \textcolor{d-green}{-42.1\%} & \scriptsize \textcolor{d-green}{-51.0\%} & 
                                \scriptsize \textcolor{d-green}{-5.56\%}
                                &
                                \scriptsize \textcolor{d-green}{-4.80\%} \\ 
\cmidrule(lr){2-12}
                                & IGEVStereo \cite{xu2023iterative}      & 0.020 & 0.020 & 0.914 & 0.761 & 0.429 & 1.15 & 0.35 & 0.100 & 0.208 & 2.36  \\
                                & {I. + BiDAStabilizer (ours)} & 0.007	& 0.007 & 0.976 & 0.882 & 0.504 & 0.62 & 0.22 & 0.050 & 0.190 & 2.18 \\
                                &    & \scriptsize \textcolor{d-green}{-65.0\%} & \scriptsize \textcolor{d-green}{-65.0\%} & \scriptsize \textcolor{d-green}{+6.78\%} & \scriptsize \textcolor{d-green}{+15.9\%} & \scriptsize \textcolor{d-green}{+17.5\%} & \scriptsize \textcolor{d-green}{-46.1\%} & \scriptsize \textcolor{d-green}{-37.1\%} & \scriptsize \textcolor{d-green}{-50.0\%} & 
                                \scriptsize \textcolor{d-green}{-8.65\%}
                                &
                                \scriptsize \textcolor{d-green}{-7.63\%} \\ 
\cmidrule(lr){2-12}
                                & DynamicStereo \cite{karaev2023dynamicstereo} & 0.016	& 0.016 & 0.927 & 0.740 & 0.375 & 0.78	& 0.23 & 0.084 & 0.248 & 3.68 \\
                                & {BiDAStereo (ours)}  & 0.013 & 0.013 & 0.929 & 0.703 & 0.434 & 0.71	& 0.24 & 0.088 & 0.310 & 4.48 \\
                                &    & \scriptsize \textcolor{d-green}{-18.8\%} & \scriptsize \textcolor{d-green}{-18.8\%} & \scriptsize \textcolor{d-green}{+0.22\%} & \scriptsize \textcolor{d-red}{-5.00\%} & \scriptsize \textcolor{d-green}{+15.7\%} & \scriptsize \textcolor{d-green}{-8.97\%} & \scriptsize \textcolor{d-red}{+4.35\%} & \scriptsize \textcolor{d-red}{+4.76\%} & 
                                \scriptsize \textcolor{d-red}{+25.0\%}
                                & \scriptsize \textcolor{d-red}{+21.7\%} \\ 
\midrule
\midrule
\rowcolor{mygray}
Training &    & \multicolumn{10}{c}{KITTI Depth}  \\
\rowcolor{mygray}
Data & \multirow{-2}{*}{Method} & OPW\(_{100}\)$\downarrow $  & OPW\(_{30}\)$\downarrow $ & RTC$\uparrow $ & TCC$\uparrow $ & TCM$\uparrow $ & $\delta_{1px}^{t}$$\downarrow $ & $\delta_{3px}^{t}$$\downarrow $ & TEPE$\downarrow $ & EPE$\downarrow $ & $\delta_{1px}$$\downarrow $ \\
\midrule
\multirow{9}{*}{\makecell{SF + DR \\ + K + ISV}}        & RAFTStereo \cite{lipson2021raft}     & 6.540 & 5.562 & 0.033 & 0.866 & 0.967 & 8.34 & 0.63 & 0.43 & 0.53 & 9.62 \\
                                & {R. + BiDAStabilizer (ours)} & 6.522 & 5.554 & 0.033 & 0.873 & 0.968 & 7.68 & 0.59 & 0.39 & 0.52 & 9.23 \\
                                &    & \scriptsize \textcolor{d-green}{-0.28\%} & \scriptsize \textcolor{d-green}{-0.14\%} & \scriptsize \textcolor{gray}{0.00\%} & \scriptsize \textcolor{d-green}{+0.81\%} & \scriptsize \textcolor{d-green}{+0.10\%} & \scriptsize \textcolor{d-green}{-7.91\%} & \scriptsize \textcolor{d-green}{-6.35\%} & \scriptsize \textcolor{d-green}{-9.30\%} & 
                                \scriptsize \textcolor{d-green}{-1.89\%}
                                & \scriptsize \textcolor{d-green}{-4.05\%} \\ 
\cmidrule(lr){2-12}
                                & IGEVStereo \cite{xu2023iterative}      & 6.634 & 5.617 & 0.033 & 0.866 & 0.966 & 8.44 & 0.66 & 0.43 & 0.54 & 9.74  \\
                                & {I. + BiDAStabilizer (ours)}  & 6.536	& 5.559 & 0.033 & 0.872 & 0.969 & 7.74 & 0.62 & 0.40 & 0.52 & 9.04 \\
                                &    & \scriptsize \textcolor{d-green}{-1.48\%} & \scriptsize \textcolor{d-green}{-1.03\%} & \scriptsize \textcolor{gray}{0.00\%} & \scriptsize \textcolor{d-green}{+0.69\%} & \scriptsize \textcolor{d-green}{+0.31\%} & \scriptsize \textcolor{d-green}{-8.29\%} & \scriptsize \textcolor{d-green}{-6.06\%} & \scriptsize \textcolor{d-green}{-6.98\%} & 
                                \scriptsize \textcolor{d-green}{-3.70\%}
                                & \scriptsize \textcolor{d-green}{-7.19\%} \\ 
\cmidrule(lr){2-12}
                                & DynamicStereo \cite{karaev2023dynamicstereo} & 6.593 & 5.589 & 0.033 & 0.866 & 0.972 & 8.21 & 0.66 & 0.42 & 0.57 & 11.17  \\
                                & {BiDAStereo (ours)}  & 6.460 & 5.544 & 0.034 & 0.865 & 0.972 & 8.33	& 0.65 & 0.42 & 0.58 & 11.38 \\
                                &    & \scriptsize \textcolor{d-green}{-2.02\%} & \scriptsize \textcolor{d-green}{-0.81\%} & \scriptsize \textcolor{d-green}{+3.03\%} & \scriptsize \textcolor{d-red}{-0.12\%} & \scriptsize \textcolor{gray}{0.00\%} & \scriptsize \textcolor{d-red}{+1.46\%} & \scriptsize \textcolor{d-green}{-1.52\%} & \scriptsize \textcolor{gray}{0.00\%} & 
                                \scriptsize \textcolor{d-red}{+1.75\%}
                                & \scriptsize \textcolor{d-red}{+1.88\%} \\ 
                                
\midrule
\midrule
\rowcolor{mygray}
Training &    & \multicolumn{10}{c}{Infinigen Stereo Video}  \\
\rowcolor{mygray}
Data & \multirow{-2}{*}{Method} & OPW\(_{100}\)$\downarrow $  & OPW\(_{30}\)$\downarrow $ & RTC$\uparrow $ & TCC$\uparrow $ & TCM$\uparrow $ & $\delta_{1px}^{t}$$\downarrow $ & $\delta_{3px}^{t}$$\downarrow $ & TEPE$\downarrow $ & EPE$\downarrow $ & $\delta_{3px}$$\downarrow $ \\
\midrule
\multirow{9}{*}{\makecell{SF + DR \\ + K + ISV}}  & RAFTStereo \cite{lipson2021raft}     & 2.161 & 1.943 & 0.649 & 0.261 & 0.250 & 17.03 & 9.42 & 2.08 & 2.84 & 12.31  \\
                                & {R. + BiDAStabilizer (ours)} & 1.455 & 1.128 & 0.786 & 0.318 & 0.301 & 14.78 & 8.28 & 1.79 & 2.74 & 12.20 \\
                                &    & \scriptsize \textcolor{d-green}{-32.7\%} & \scriptsize \textcolor{d-green}{-41.9\%} & \scriptsize \textcolor{d-green}{+21.1\%} & \scriptsize \textcolor{d-green}{+21.8\%} & \scriptsize \textcolor{d-green}{+20.4\%} & \scriptsize \textcolor{d-green}{-13.2\%} & \scriptsize \textcolor{d-green}{-12.1\%} & \scriptsize \textcolor{d-green}{-13.9\%} &
                                \scriptsize \textcolor{d-green}{-3.52\%}
                                &
                                \scriptsize \textcolor{d-green}{-0.89\%} \\ 
\cmidrule(lr){2-12}
                                & IGEVStereo \cite{xu2023iterative}      & 11.300 & 4.720 & 0.633 & 0.277 & 0.226 & 16.65 & 9.04 & 1.96 & 2.45 & 10.86  \\
                                & {I. + BiDAStabilizer (ours)} & 1.653	& 1.354 & 0.772 & 0.338 & 0.269 & 14.82 & 8.31 & 1.69 & 2.42 & 10.94  \\
                                &    & \scriptsize \textcolor{d-green}{-85.4\%} & \scriptsize \textcolor{d-green}{-71.3\%} & \scriptsize \textcolor{d-green}{+22.0\%} & \scriptsize \textcolor{d-green}{+22.0\%} & \scriptsize \textcolor{d-green}{+19.0\%} & \scriptsize \textcolor{d-green}{-11.0\%} & \scriptsize \textcolor{d-green}{-8.07\%} & \scriptsize \textcolor{d-green}{-13.8\%} & 
                                \scriptsize \textcolor{d-green}{-1.22\%}
                                &
                                \scriptsize \textcolor{d-red}{+0.74\%} \\ 
\cmidrule(lr){2-12}
                                & DynamicStereo \cite{karaev2023dynamicstereo} & 1.313 & 1.777 & 0.717 & 0.274 & 0.228 & 15.83 & 8.65 & 1.92 & 4.13 & 15.17  \\
                                & {BiDAStereo (ours)}   & 1.070 & 1.037 & 0.692 & 0.268 & 0.231 & 15.20 & 8.37 & 1.89 & 3.47 & 13.56  \\
                                &    & \scriptsize \textcolor{d-green}{-18.5\%} & \scriptsize \textcolor{d-green}{-41.6\%} & \scriptsize \textcolor{d-red}{-3.49\%} & \scriptsize \textcolor{d-red}{-2.19\%} & \scriptsize \textcolor{d-green}{+1.32\%} & \scriptsize \textcolor{d-green}{-3.98\%} & \scriptsize \textcolor{d-green}{-3.24\%} & \scriptsize \textcolor{d-green}{-1.56\%}  & 
                                \scriptsize \textcolor{d-green}{-16.0\%}
                                & \scriptsize \textcolor{d-green}{-10.6\%} \\ 

\bottomrule[1.0pt]
\end{tabular}
\label{tab: robust}
\end{table*}

\subsection{Metrics}
To evaluate spatial accuracy, we use the standard metrics: end-point-error (EPE) and $\delta_{n-px}$, the proportion of pixels with EPE greater than the $n$ threshold. For temporal consistency, we follow \cite{karaev2023dynamicstereo} and use the temporal end-point-error (TEPE), which measures the variation of the EPE across the temporal dimension:
\begin{equation}
\text{TEPE} = \sum_{t=1}^{T-1} \left\lVert ((\mathbf{d}^{t} - \mathbf{d}^{t+1}) - (\mathbf{d}_{\mathrm{gt}}^{t} - \mathbf{d}_{\mathrm{gt}}^{t+1})) \right\rVert_2,
\end{equation}
$\delta_{n-px}^{t}$ represents the proportion of pixels with TEPE exceeding $n$. Lower values across all metrics indicate better temporal consistency. Additionally, we use the optical flow-based warping error (OPW) metric \cite{lai2018learning,cao2021learning,wang2022less}, which evaluates the change in a pixel’s depth across adjacent frames, formulated as:
\begin{equation}
\text{OPW} = \sum_{t=1}^{T-1} \left\lVert \mathbf{O}^t \cdot \left(\hat{\mathbf{D}}^{t+1} - \mathbf{D}^t\right) \right\rVert,
\end{equation}
where $\mathbf{O}^t=\exp(-50 \cdot(\hat{\mathbf{I}}^{t+1}_L - \mathbf{I}^t_L))$ is an occlusion mask, and $\mathbf{D}$ is the real depth, converted from disparity. OPW represents the average meter change in the depth of a point across frames. $\text{OPW}_{n}$ denotes the metric calculated for pixels within an $n$ meter range. To better assess human perception of consistency, we follow \cite{li2021enforcing} and calculate the Relative Temporal Consistency (RTC) metric:
\begin{equation}
\text{RTC} = \sum_{t=1}^{T-1} \left\lVert \mathbf{1} \left( \mathbf{M}^t \cdot \max \left( \frac{\hat{\mathbf{D}}^{t+1}}{\mathbf{D}^t}, \frac{\mathbf{D}^t}{\hat{\mathbf{D}}^{t+1}} \right) < 1.01 \right) \right\rVert,
\end{equation}
where $\mathbf{1}(\cdot)$ is the indicator function for the set of pixels that satisfy the enclosed inequality. Additionally, we utilize the temporal change consistency (TCC) and the temporal motion consistency (TCM) metrics as proposed in \cite{khan2023temporally}:
\begin{equation}
    \begin{aligned}
    \text{TCC} &= \sum_{t=1}^{T-1} \text{SSIM} \left( \left| \hat{\mathbf{D}}^{t+1} - \mathbf{D}^t \right|, \left| \hat{\mathbf{D}}_{gt}^{t+1} - \mathbf{D}_{gt}^t \right| \right),\\
    \text{TCM} &= \sum_{t=1}^{T-1} \text{SSIM} \left( F\left( \mathbf{D}^t, \hat{\mathbf{D}}^{t+1} \right), F\left( \hat{\mathbf{D}}_{gt}^{t+1} - \mathbf{D}_{gt}^t \right) \right).
    \end{aligned}
\end{equation}
Here, SSIM \cite{wang2004image} refers to the structural similarity index, and $F(\cdot)$ is optical flow estimation from RAFT \cite{teed2020raft}.

\subsection{Implementation Details} \label{Implementation Details}
\textcolor{black}{We implemented BiDAStabilizer and BiDAStereo in PyTorch and trained them on NVIDIA A100 GPUs. RAFT \cite{teed2020raft} was used as the optical flow module. For BiDAStabilizer, the entire training process was set to 8k iterations with a resolution of $256 \times 512$. For BiDAStereo, we first pre-trained the model with a batch size of 16 and a resolution of $256 \times 256$, followed by fine-tuning with a batch size of 8 and a resolution of $256 \times 512$. The pre-training process consisted of 60k iterations for the SF version and 80k iterations for the combined datasets version. Fine-tuning included 40k iterations for the SF version and 60k iterations for the combined datasets version. Training took approximately 0.5 days for BiDAStabilizer and 5 days for BiDAStereo. We used the AdamW optimizer \cite{adam} with a standard learning rate of 0.0004 and default settings, along with the one-cycle learning rate schedule \cite{smith2019super}. Following \cite{karaev2023dynamicstereo}, multiple data augmentation techniques were applied during training, including random cropping, rescaling, and saturation shifts. The sequence length was set to $T=5$ during training for BiDAStereo and $T=8$ for BiDAStabilizer. During inference, we set $T=20$ for the DR and ISV datasets, $T=40$ for KITTI, and $T=50$ for Sintel. The number of iterations was set to 10 during training and 20 during evaluation for BiDAStereo.}

\begin{figure}[t]
   \begin{center}
   \includegraphics[width=1\linewidth]{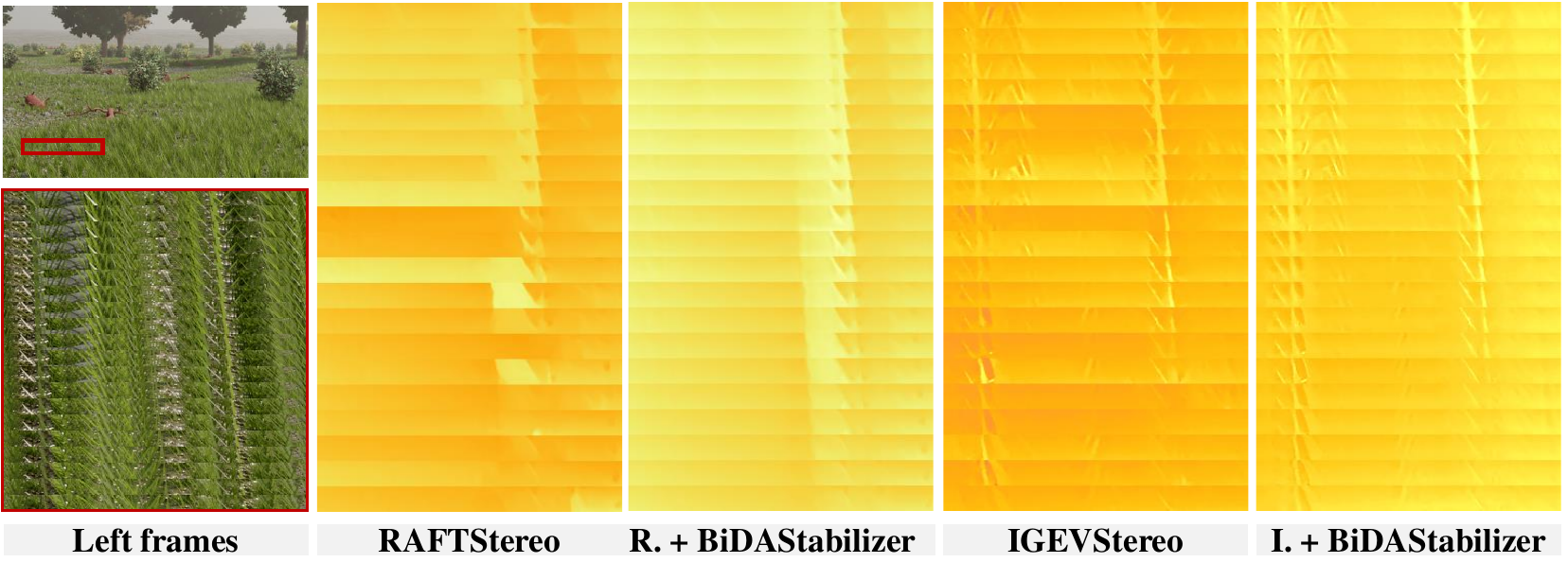}
   \end{center}
   \vspace{-.5em}
      \caption{Qualitative comparisons on the Infinigen SV test set. Our BiDAStabilizer ensures flicker-free disparity estimation. The top-left image shows a sample frame, while the subsequent images depict scanline slices (20 frames) through the spatio-temporal domain across different methods.
}
   \label{fig:infinigen-comparison}
\end{figure}

\begin{figure}[t]
   \begin{center}
   \includegraphics[width=1\linewidth]{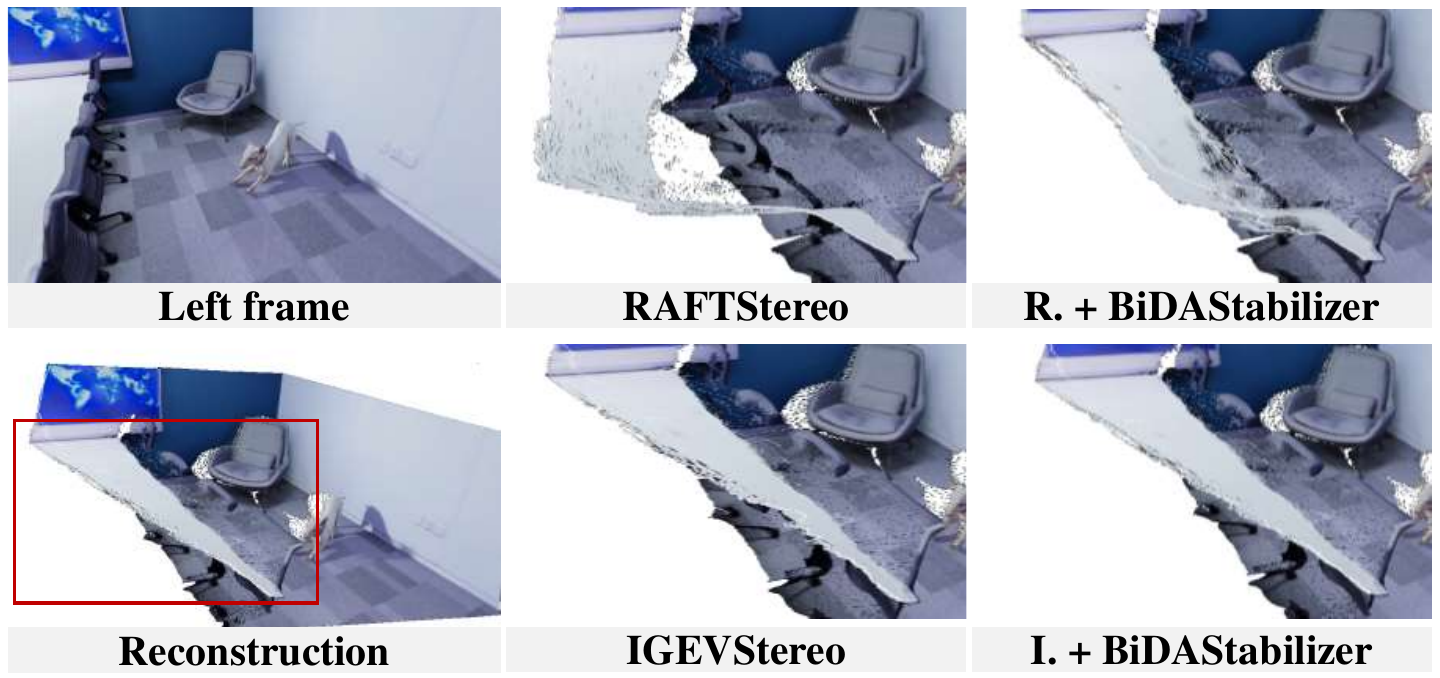}
   \end{center}
    \vspace{-.5em}
      \caption{Qualitative comparisons on the Dynamic Replica test set \cite{karaev2023dynamicstereo}. Disparities are converted to globally aligned point clouds and rendered via a camera, visualized for comparison.}
   \label{fig:dr-comparison}
\end{figure}

\begin{figure*}[ht]
  \centering
  \includegraphics[width=1\textwidth]{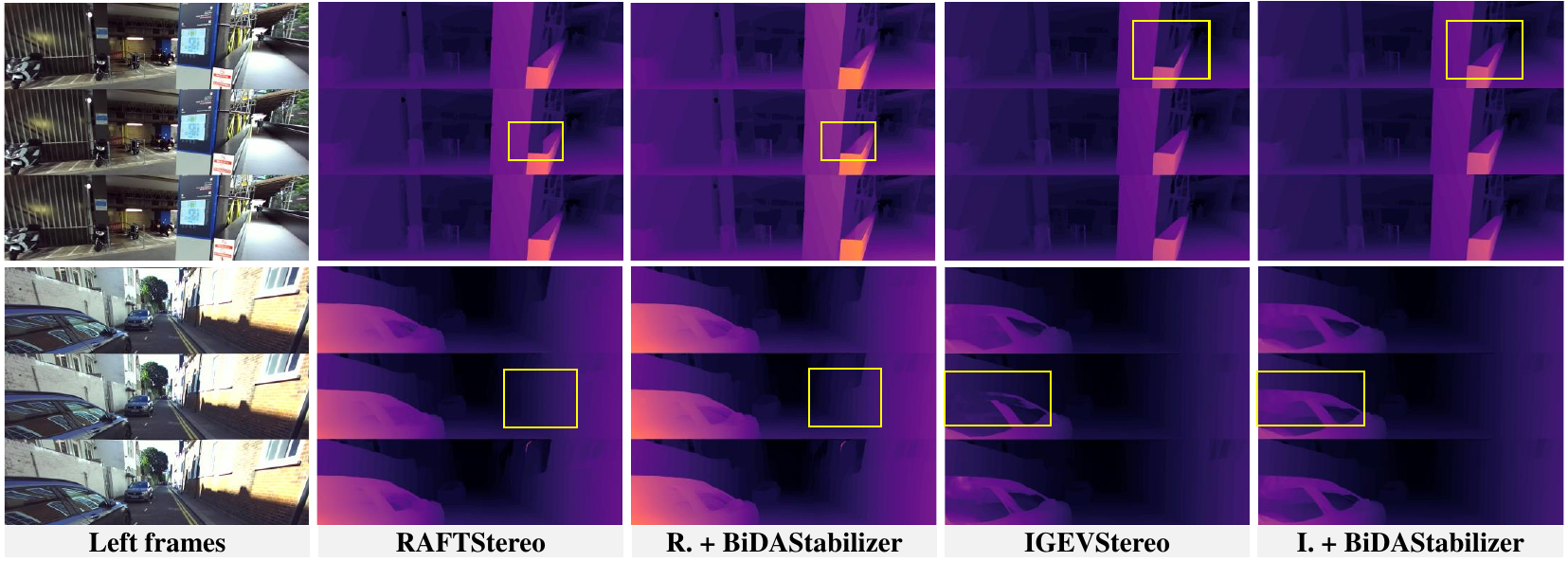}
     \vspace{-.9em}
  \caption{Temporal consistency comparisons on outdoor scenarios from the South Kensington SV Dataset. Our BiDAStabilizer reduces temporal inconsistencies, resulting in improved frame-to-frame consistency, while preserving the spatial quality of the disparity.}
  \label{fig:sk-outdoor-comparison}
\end{figure*}

\begin{figure*}[ht]
  \centering
  \includegraphics[width=1\textwidth]{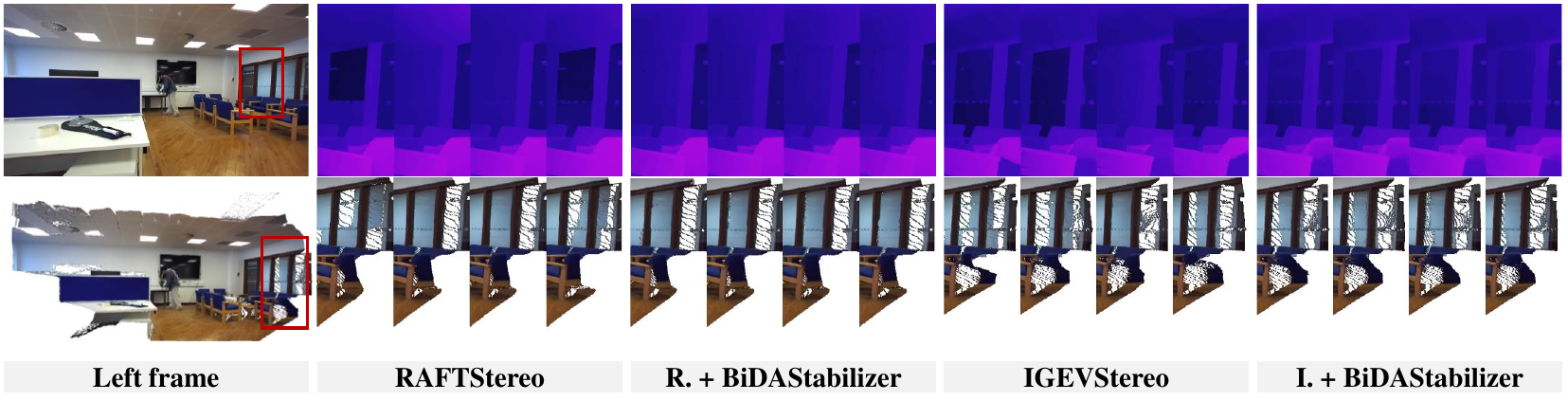}
  \vspace{-.9em}
  \caption{Temporal consistency comparisons on a dynamic indoor scenario from the South Kensington SV Dataset. \textbf{Top:} Input left frames and predicted disparities for the same region across four different frames, using various methods. \textbf{Bottom:} Disparities are converted to globally aligned point clouds and rendered via a camera, visualized for comparison. Our BiDAStabilizer produces consistent and accurate disparities without flickering.}
  \label{fig:sk-indoor-comparison}
\end{figure*}

\subsection{Temporal Consistency}

\textbf{In-domain evaluation.}
Tab.~\ref{tab:in domain} presents a comparison of the numerical results obtained by our plugin method, BiDAStabilizer, and our video method, BiDAStereo, against other comparison methods across the Dynamic Replica, KITTI Depth, and Infinigen SV datasets. When trained and evaluated within the source domain, our methods demonstrate superior temporal consistency. Notably, BiDAStabilizer achieves significant improvements in TEPE, with gains of 43.9\% and 43.4\% on Dynamic Replica, 5.56\% and 7.89\% on KITTI Depth, and 14.7\% and 14.8\% on Infinigen SV for RAFTStereo \cite{lipson2021raft} and IGEVStereo \cite{xu2023iterative}, respectively. Similar performance advantages are observed in OPW, RTC, TCC, TCM, and $\delta^{t}$ metrics. From this table we can also see that our video method BiDAStereo outperforms DynamicStereo among most of the metrics, underscoring the effectiveness of leveraging temporal information in our approaches.

\textbf{Out-of-domain evaluation.} 
\textcolor{black}{Tab.~\ref{tab:out of domain} demonstrates the results when models are trained on the source domain and evaluated on the target domain (Sintel dataset). The results clearly show that our methods achieve superior performance across nearly all metrics, both in terms of accuracy and temporal consistency. Specifically, when trained solely on SceneFlow, the proposed BiDAStabilizer enhances RAFTStereo \cite{lipson2021raft} by 6.52\% and 8.10\% in TEPE on the Sintel clean and final passes, respectively. When trained on both SceneFlow and Dynamic Replica, BiDAStabilizer further outperforms the original RAFTStereo, with improvements of 11.2\% and 9.42\% in TEPE for the clean and final passes. Similar performance gains are observed for IGEVStereo \cite{xu2023iterative} and across other evaluation metrics. Moreover, our BiDAStereo surpasses DynamicStereo \cite{karaev2023dynamicstereo}, showing a 53.3\% and 37.9\% improvement in EPE on the final pass compared to DynamicStereo. These results underscore the strong generalization capability of our methods when applied to unseen domains.}

\begin{table*}[tb]
\footnotesize
   \begin{center}
       \caption{Generalization evaluation from different training datasets to Sintel \cite{sintel} for RAFTStereo \cite{lipson2021raft} and BiDAStereo. SF - SceneFlow \cite{mayer2016large}, DR - Dynamic Replica \cite{karaev2023dynamicstereo}, ISV - proposed Infinigen Stereo Video.}
\vspace{-.5em}
     \begin{tabular}{m{2.0cm}<{\centering}|m{1.05cm}<{\centering}m{1.05cm}<{\centering}m{1.05cm}<{\centering}m{1.05cm}<{\centering}m{1.05cm}<{\centering}|m{1.05cm}<{\centering}m{1.05cm}<{\centering}m{1.05cm}<{\centering}m{1.05cm}<{\centering}m{1.05cm}<{\centering}}
        \toprule[1.0pt]
        Method & \multicolumn{5}{c}{RAFTStereo (Image based)} & \multicolumn{5}{|c}{BiDAStereo (Video based)} \\
        \midrule
        \rowcolor{mygray}
         Training & \multicolumn{10}{c}{Sintel Clean}  \\
        \rowcolor{mygray}
        {Data} & $\delta_{1px}^{t}$$\downarrow $ & $\delta_{3px}^{t}$$\downarrow $ & TEPE$\downarrow $  & $\delta_{3px}$$\downarrow $ & EPE$\downarrow $ & $\delta_{1px}^{t}$$\downarrow $ & $\delta_{3px}^{t}$$\downarrow $ & TEPE$\downarrow $  & $\delta_{3px}$$\downarrow $ & EPE$\downarrow $ \\
        \midrule
        SF     & 9.32 & 4.51 & 0.92 & 6.12 & 1.44 & 8.29 & 3.79 & \textbf{0.73} & 5.94 & 1.26 \\
        SF + DR  & 9.07  & 4.40 & 0.89 & 5.83 & \textbf{1.20} & 8.03 & 3.76 & 0.75 & 5.75 & 1.30 \\
        SF + ISV (ours) & \textbf{8.81}	& \textbf{4.20} & \textbf{0.85} & \textbf{5.69} & \textbf{1.34} & \textbf{7.93} & \textbf{3.66} & \textbf{0.73} & \textbf{5.64} & \textbf{1.23} \\
        \midrule
        \midrule
        \rowcolor{mygray}
        Training & \multicolumn{10}{c}{Sintel Final}  \\
        \rowcolor{mygray}
        Data & $\delta_{1px}^{t}$$\downarrow $ & $\delta_{3px}^{t}$$\downarrow $ & TEPE$\downarrow $  & $\delta_{3px}$$\downarrow $ & EPE$\downarrow $ & $\delta_{1px}^{t}$$\downarrow $ & $\delta_{3px}^{t}$$\downarrow $ & TEPE$\downarrow $  & $\delta_{3px}$$\downarrow $ & EPE$\downarrow $ \\
        \midrule
        SF    & 13.69 & 7.08 & 2.10 & 10.39 & 4.89 & 11.65 & 5.53 & 1.26 & 8.78 & 1.99  \\
        SF + DR  & 13.56 & 7.02 & 1.91 & 9.83 & 4.52 & {11.04} & \textbf{5.30} & \textbf{1.22} & \textbf{8.52} & 1.98 \\
        SF + ISV (ours) & \textbf{12.77} & \textbf{6.45} & \textbf{1.75} & \textbf{9.46} & \textbf{4.49} & \textbf{11.03} & 5.33 & 1.23 & 8.70 & \textbf{1.97} \\
        \bottomrule[1.0pt]
     \end{tabular}
    \label{tab:ablation dataset}   
   \end{center}
\end{table*}

\begin{table*}[t]
\caption{Ablation Study on iterative updater, alignment, propagation components of BiDAStereo on Sintel \cite{sintel} and Dynamic Replica \cite{karaev2023dynamicstereo}. Lower values indicate better results for all metrics. The approach used in our final model is underlined. See Sec.~\ref{Ablation Studies} for details.}
\vspace{-.5em}
\setlength{\tabcolsep}{8.pt}
\footnotesize
    \begin{center}
        \begin{tabular}{cc|cccc|cccc|c}
        \toprule[1.0pt]
        \rowcolor{mygray}
         &  & \multicolumn{4}{c|}{Sintel Final}   & \multicolumn{4}{c|}{Dynamic Replica} &   \\
        \rowcolor{mygray}
         \multirow{-2}{*}{Component} & \multirow{-2}{*}{Setup} & $\delta_{1px}^{t}$$\downarrow $ & $\delta_{3px}^{t}$$\downarrow $ & TEPE$\downarrow $ & $\delta_{3px}$$\downarrow $ & $\delta_{1px}^{t}$$\downarrow $ & $\delta_{3px}^{t}$$\downarrow $ & TEPE$\downarrow $ & $\delta_{1px}$$\downarrow $ & \multirow{-2}{*}{Parameters} \\
        \midrule
        \multirow{4}{*}{Updater} & 2D Conv, $1\times 5$ & 12.83 & 6.50 & 1.74 & 8.45 &	0.84 & 0.27 & 0.091 & 2.40 & {5.0M}          \\
        & 3D Conv, $1\times 1 \times 5$ & 11.22 & 5.61 & 1.34 &  8.64 & 0.69 & 0.23 & 0.082 & 3.38  & 5.4M \\
        & 3D Conv, $1\times 3 \times 3$ & 13.41 & 5.99 & 1.43 & 8.93  & 0.91  & 0.27 & 0.087 & 3.48 & 6.3M \\
        & \underline{3D Conv, $1\times 1 \times 15$} & \textbf{11.19} & \textbf{5.56} & \textbf{1.33} & \textbf{8.59} & \textbf{0.69} & \textbf{0.23} & \textbf{0.079} & \textbf{3.26} & 6.9M \\
        \midrule
        % \multirow{3}{*}{Alignment} & w/o Alignment & 11.19 & 5.56 & 1.33 & 8.59 & 0.69 & 0.23 & 0.079 & 3.26 & \textbf{6.9M} \\
        \multirow{2}{*}{Alignment} & \underline{Single-Multi} & \textbf{11.10} & \textbf{5.38} & \textbf{1.31} & \textbf{8.53}  & \textbf{0.64} & \textbf{0.22} & \textbf{0.067} & \textbf{2.91} & 12.2M \\
        & Multi-Multi & 11.30 & 5.77 & 1.39 & 8.66  & 0.74 & 0.25 & 0.082 & 3.28 & 12.2M \\
        % \cmidrule(lr){2-11}
        % & SpyNet  &   &   &  &    &   &   &   &   &   \\
        % & RAFT  &   &   &  &    &   &   &   &   &   \\
        \midrule
        % \multirow{4}{*}{Propagation} & w/o Propagation & 11.10 & 5.38 & 1.31 & 8.53  & 0.64 & \textbf{0.22} & 0.067 & 2.91 & 12.2M \\
        \multirow{3}{*}{Propagation} & Motion, Separated & 20.36 & 12.08 & 2.82 & 20.38  & 3.84 & 2.25 & 0.511 & 6.07 &  12.2M   \\
        & \underline{Motion, Shared} & \textbf{11.09} & \textbf{5.33} & \textbf{1.24} & 8.55 & \textbf{0.61} & \textbf{0.22} & \textbf{0.062} & 2.81 & 12.2M \\
        & Feature, Shared   & 12.39 & 6.25 & 1.53  & \textbf{8.23} &  0.77  & 0.27 & 0.100   & \textbf{2.19}  & 11.8M \\
        
        \bottomrule[1.0pt]
        \end{tabular}
    \end{center}
\label{tab:ablation}
\end{table*}

\textbf{Robustness evaluation.} \textcolor{black}{Inspired by the Robust Vision Challenge \cite{robustvision}, we perform a robustness evaluation by training models on mixed datasets and testing them across different benchmarks using a fixed set of parameters. As shown in Tab.~\ref{tab: robust}, our BiDAStabilizer achieves the best results in terms of temporal consistency. Specifically, it improves the TEPE of RAFTStereo by 51.0\% on Dynamic Replica, 9.30\% on KITTI Depth, and 13.9\% on Infinigen SV. Similar gains are observed for IGEVStereo, with improvements of 50.0\%, 6.98\%, and 13.8\% across the same benchmarks, respectively. Notably, our method also enhances overall accuracy, further demonstrating its effectiveness. In addition, our video method also surpasses existing SoTA DynamicStereo \cite{karaev2023dynamicstereo} on most of the metrics. Figs.~\ref{fig:infinigen-comparison} and \ref{fig:dr-comparison} provide qualitative comparisons on the Infinigen SV and Dynamic Replica test sets, highlighting the improvement in temporal consistency. Our plugin network significantly reduces flicker in disparity estimation, enhancing the temporal consistency of RAFTStereo and IGEVStereo.}

\subsection{Ablation Studies} \label{Ablation Studies}

\begin{table}[t]
\footnotesize
   \begin{center}
       \caption{\textcolor{black}{Ablation study on temporal aggregation of BiDAStabilizer.}}
       \vspace{-.5em}
     \begin{tabular}{m{2.5cm}<{\centering}|m{0.95cm}<{\centering}m{0.95cm}<{\centering}|m{0.95cm}<{\centering}m{0.95cm}<{\centering}}
        \toprule[1.0pt]
        \rowcolor{mygray}
         & \multicolumn{2}{c}{Sintel Final}      & \multicolumn{2}{c}{Dynamic Replica} \\
        \rowcolor{mygray}
         \multirow{-2}{*}{Setup} & $\delta_{1px}^{t}$$\downarrow $ & TEPE$\downarrow $ & $\delta_{1px}^{t}$$\downarrow $  & TEPE$\downarrow $\\
         \midrule
        3D UNet & 21.21 & 2.23 & 7.99 & 0.358 \\
        \midrule
        3D Conv & 14.69 & 1.96 & 1.26 & 0.117 \\
        w. 1D attention & 14.83 & 2.03 & 1.25 & 0.116 \\
        w. 1D + 2D attention & 14.51 & 1.97 & 0.94 & 0.088 \\
        w. 3D attention & \multicolumn{4}{c}{Out of Memory} \\
        \midrule
        Propogation & \textbf{12.31} & \textbf{1.73} & \textbf{0.60}  & \textbf{0.046} \\
        \bottomrule[1.0pt]
     \end{tabular}
     \label{tab:ablation bidastabilizer attention}
   \end{center}
\end{table}

\textbf{Real-world scenes.} \textcolor{black}{We visualize the results of different methods on our proposed South Kensington SV dataset. Fig.~\ref{fig:sk-outdoor-comparison} presents the outcomes on outdoor scenes. While RAFTStereo \cite{lipson2021raft} and IGEVStereo \cite{xu2023iterative} perform well on image-based benchmarks, they struggle to produce consistent disparities in real-world stereo videos. In contrast, the disparities produced with our BiDAStabilizer are significantly more consistent. Fig.~\ref{fig:sk-indoor-comparison} shows the visualization results for dynamic indoor scenes. Images are converted to point clouds using depth predictions and the reconstruction images are rendered with the camera positioned at the first frame pose. As seen, our method notably improves temporal consistency, particularly in challenging areas like transparent windows, resulting in a more stable and consistent reconstruction quality.}

\textcolor{black}{\textcolor{black}{In Tabs.~\ref{tab:ablation dataset}, \ref{tab:ablation}, \ref{tab:ablation bidastabilizer}, \ref{tab:ablation bidastabilizer attention}, \ref{tab:ablation bidastereo}, and \ref{tab:ablation inference}, we evaluated the proposed dataset and specific components of our approaches in isolation and highlight the settings used in the final model.} The models are trained on the SceneFlow \cite{mayer2016large} and Dynamic Replica \cite{karaev2023dynamicstereo}, using the hyperparameters detailed in Sec.~\ref{Implementation Details}. We then evaluate these models on the final pass of Sintel \cite{sintel} and on the test split of Dynamic Replica \cite{karaev2023dynamicstereo}.}

\textbf{Infinigen SV dataset.} \textcolor{black}{Tab.~\ref{tab:ablation dataset} presents the results of training a state-of-the-art image-based model, RAFTStereo \cite{lipson2021raft}, and our video-based model, BiDAStereo, on various combinations of SceneFlow, Dynamic Replica, and the proposed Infinigen SV datasets to evaluate their generalization ability on Sintel \cite{sintel}. The results indicate that incorporating Infinigen SV alongside SceneFlow enhances both models' performance in terms of temporal consistency and accuracy. This demonstrates that the value of Infinigen SV extends beyond improving temporal consistency in video-based training, offering benefits in standard image-based training as well.}

\textbf{BiDAStabilizer components.} \textcolor{black}{In Tab.~\ref{tab:ablation bidastabilizer}, we analyze the individual components of BiDAStabilizer. The results show that removing the alignment operation, propagation mechanism, or feature-based fusion leads to varying degrees of performance degradation. This highlights the importance and effectiveness of each component in the overall performance of BiDAStabilizer.} \textcolor{black}{In Tab.~\ref{tab:ablation bidastabilizer attention}, we replace the propagation module in BiDAStabilizer with 3D convolution–based architectures and attention variants, while keeping the feature extraction part unchanged. As shown, our proposed propagation consistently outperforms these baselines on both benchmarks, demonstrating its superior ability to model temporal consistency.}

\textbf{Updater.} \textcolor{black}{As shown in Tab.~\ref{tab:ablation}, we compare 2D and 3D convolutions in the iterative updater (Sec.~\ref{Motion-propagation based Recurrent Unit}). The table reveals that, compared to the 2D model, employing 3D convolutions significantly enhances temporal consistency, with a notable improvement in TEPE from 1.74 to 1.34. This demonstrates the effectiveness of processing multiple frames simultaneously. However, it is also observed that the accuracy of the method with 3D convolutions decreases slightly, likely due to the fusion of unaligned features across time. While prior work \cite{karaev2023dynamicstereo} utilized $1\times1\times5$ convolutions in the updater, we found that enlarging the kernel size across horizontal dimensions leads to a general improvement in both accuracy and temporal consistency.}

\textbf{Alignment.} \textcolor{black}{After incorporating bidirectional alignment into the correlation layer, we observed significant performance improvements across all metrics. We also experimented with constructing triple-frame cost volumes for both adjacent aligned left frames and their corresponding aligned right frames (multi-multi). However, this approach resulted in a performance drop, likely due to the additional noise introduced by using more than one reference image. For pixel-level matching, using a fixed reference image, along with the expanded temporal receptive field of the source images, yielded better results.}

\begin{table}[t]
\footnotesize
   \begin{center}
       \caption{Ablation study on components of BiDAStabilizer.}
       \vspace{-.5em}
     \begin{tabular}{m{2.5cm}<{\centering}|m{0.95cm}<{\centering}m{0.95cm}<{\centering}|m{0.95cm}<{\centering}m{0.95cm}<{\centering}}
        \toprule[1.0pt]
        \rowcolor{mygray}
         & \multicolumn{2}{c}{Sintel Final}      & \multicolumn{2}{c}{Dynamic Replica} \\
        \rowcolor{mygray}
         \multirow{-2}{*}{Setup} & $\delta_{1px}^{t}$$\downarrow $ & TEPE$\downarrow $ & $\delta_{1px}^{t}$$\downarrow $  & TEPE$\downarrow $\\
        \midrule
        Full & \textbf{12.31} & \textbf{1.73} & \textbf{0.60}  & \textbf{0.046} \\
        w/o Alignment & 13.43  & 1.88   & 0.81  & 0.070 \\
        w/o Propogation & 13.18  & 1.85   & 0.79  & 0.072    \\
        w/o Feature-based & 13.10  & 1.79   & 0.67  & 0.055    \\
        \bottomrule[1.0pt]
     \end{tabular}
     \label{tab:ablation bidastabilizer}
   \end{center}
\end{table}

\begin{table}[t]
\footnotesize
   \begin{center}
       \caption{Ablation study on the proposed local matching and global propagation of BiDAStereo.}
       \vspace{-.5em}
     \begin{tabular}{m{.9cm}<{\centering}m{.9cm}<{\centering}|m{1.1cm}<{\centering}m{1.1cm}<{\centering}|m{1.1cm}<{\centering}m{1.1cm}<{\centering}}
        \toprule[1.0pt]
        \rowcolor{mygray}
        \multicolumn{2}{c|}{Setup} & \multicolumn{2}{c|}{Sintel Final}      & \multicolumn{2}{c}{Dynamic Replica} \\
        \rowcolor{mygray}
         Local& Global & $\delta_{1px}^{t}$$\downarrow $ & TEPE$\downarrow $ & $\delta_{1px}^{t}$$\downarrow $  & TEPE$\downarrow $\\
        \midrule
         \XSolidBrush & \XSolidBrush & 11.19 & 1.33   & 0.69  & 0.079   \\
         \Checkmark & \XSolidBrush & 11.10  & 1.31   & 0.64  & 0.067   \\
         \XSolidBrush & \Checkmark & 13.97   & 1.78   & 0.82  & 0.076   \\
        \Checkmark  & \Checkmark & \textbf{11.09}  & \textbf{1.24}  & \textbf{0.61}  & \textbf{0.062}   \\
        \bottomrule[1.0pt]
     \end{tabular}
     \label{tab:ablation bidastereo}
   \end{center}
\end{table}

\begin{table}[t]
\footnotesize
   \begin{center}
       \caption{Ablation study on the inference of BiDAStereo.}
       \vspace{-.5em}
     \begin{tabular}{m{1.0cm}<{\centering}m{1.0cm}<{\centering}|m{1.0cm}<{\centering}m{1.0cm}<{\centering}|m{1.0cm}<{\centering}m{1.0cm}<{\centering}}
        \toprule[1.0pt]
        \rowcolor{mygray}
         &  & \multicolumn{2}{c}{Sintel Final}      & \multicolumn{2}{c}{Dynamic Replica} \\
        \rowcolor{mygray}
        \multirow{-2}{*}{Inference} & \multirow{-2}{*}{Setup} & $\delta_{1px}^{t}$$\downarrow $ & TEPE$\downarrow $ & $\delta_{1px}^{t}$$\downarrow $  & TEPE$\downarrow $\\
        \midrule
        \multirow{2}{*}{Frame} & 12 & 11.38  & 1.32   &  0.66 & 0.081    \\
        & \underline{20} & \textbf{11.09}  & \textbf{1.24}   & \textbf{0.61}  & \textbf{0.062}    \\
        \midrule
        \multirow{3}{*}{Iteration}  & 12  & 11.30 & 1.26 & 0.61 & 0.061 \\
                                    & \underline{20}  & \textbf{11.09} & 1.24 & \textbf{0.61} & \textbf{0.062} \\
                                    & 32  & 11.10 & \textbf{1.23} & 0.62 & 0.063 \\
        \bottomrule[1.0pt]
     \end{tabular}
     \label{tab:ablation inference}
   \end{center}
\end{table}

\begin{table}[t]
\footnotesize
   \begin{center}
       \caption{\textcolor{black}{Comparison with depth prior based method trained on SF \cite{mayer2016large}.}}
       \vspace{-1em}
     \begin{tabular}{m{3.0cm}<{\centering}|m{0.8cm}<{\centering}m{0.8cm}<{\centering}|m{0.8cm}<{\centering}m{0.8cm}<{\centering}}
        \toprule[1.0pt]
        \rowcolor{mygray}
         & \multicolumn{2}{c}{Sintel Final}      & \multicolumn{2}{c}{Dynamic Replica} \\
        \rowcolor{mygray}
         \multirow{-2}{*}{Setup} & $\delta_{1px}^{t}$$\downarrow $ & TEPE$\downarrow $ & $\delta_{1px}^{t}$$\downarrow $  & TEPE$\downarrow $\\
        \midrule
        Stereo Any Video \cite{jing2025stereovideotemporallyconsistent}  &  9.98 & 1.07 & \textbf{0.67}  & \textbf{0.057}  \\
        \midrule
        BiDAStereo & 11.65 & 1.26  & 1.04 & 0.091  \\
        BiDAStereo (w. VDA) & \textbf{9.93} & \textbf{1.05} & 0.78 & 0.067 \\
        \bottomrule[1.0pt]
     \end{tabular}
     \label{tab:depth prior}
   \end{center}
\end{table}

\begin{table}[t]
\footnotesize
   \begin{center}
       \caption{\textcolor{black}{Comparison with monocular video depth methods.}}
       \vspace{-1em}
     \begin{tabular}{m{3.0cm}<{\centering}|m{0.8cm}<{\centering}m{0.8cm}<{\centering}|m{0.8cm}<{\centering}m{0.8cm}<{\centering}}
        \toprule[1.0pt]
        \rowcolor{mygray}
         & \multicolumn{2}{c}{Sintel Final}      & \multicolumn{2}{c}{Dynamic Replica} \\
        \rowcolor{mygray}
         \multirow{-2}{*}{Setup} & EPE$\downarrow $  & TEPE$\downarrow $ &  EPE$\downarrow $ & TEPE$\downarrow $  \\
         \midrule
         DepthCrafter \cite{hu2024-DepthCrafter}  & 8.90 & 2.22 & 1.36 & 0.084  \\
         VDA-S \cite{chen2025videodepthanythingconsistent}  & 10.57 & 2.85 & 1.79 & 0.077\\
         VDA-L \cite{chen2025videodepthanythingconsistent}  & 10.15 & 2.73 & 1.42 & 0.066 \\
         \midrule
         BiDAStereo & \textbf{1.98} & \textbf{1.22} & \textbf{0.22} & \textbf{0.062} \\
        \bottomrule[1.0pt]
     \end{tabular}
     \label{tab:monocular}
   \end{center}
\end{table}

\begin{table}[t]
\footnotesize
   \begin{center}
       \caption{Comparison of different methods on model trainable parameters, inference GPU memory (20 frames$\times$720$\times$1280), multiply-accumulate (MAC, 720$\times$1280), and training iterations.}
       \vspace{-.5em}
     \begin{tabular}{m{2.6cm}<{\centering}|m{1.0cm}<{\centering}m{1.0cm}<{\centering}m{1.0cm}<{\centering}m{1.2cm}<{\centering}}
        \toprule[1.0pt]
        \rowcolor{mygray}
        Methods &  Para. (M) & Mem. (G) & MACs (T) & Train. Iter (K) \\
        \midrule
        BiDAStabilizer (ours)   & 0.7  & 13.8  & 31.3  & 8  \\
        \midrule
        RAFTStereo \cite{lipson2021raft} & 11.1 & 5.4 & 78.2 & 40  \\
        IGEVStereo \cite{xu2023iterative} & 12.5 & 4.6 & 65.6 & 40  \\
        \midrule
        DynamicStereo \cite{karaev2023dynamicstereo} & 20.5 & 35.1 & 182.3 & 70 \\
        BiDAStereo (ours)  & 12.2 & 41.1 & 186.6 & 100   \\
        \bottomrule[1.0pt]
     \end{tabular}
\label{tab:efficiency}
   \end{center}
\end{table}

\textbf{Propagation.} \textcolor{black}{Motion propagation is developed to leverage the global temporal information from the whole sequences. As shown in Tab.~\ref{tab:ablation}, the motion hidden state from the previous update module was upscaled to the next one (shared), which proved to be effective. We also evaluated the approach of separating and re-initializing the motion hidden state at each resolution (separated). However, this method resulted in unstable training and a significant performance drop. Tab.~\ref{tab:ablation bidastereo} presents the ablation study on the proposed local and global modules. Interestingly, applying only global aggregation led to a performance decline, highlighting the importance of local matching as a fundamental component.}

\textbf{Inference settings.} \textcolor{black}{We also explored different iteration numbers and numbers of frames during inference. As indicated in Tab.~\ref{tab:ablation inference}, 20 iterations yielded optimal performance. Additionally, processing more frames simultaneously during inference further improved the results.}

\subsection{Adaptation with Depth Prior}
\textcolor{black}{In Tab.~\ref{tab:depth prior}, we investigate the adaptation of rich depth priors to enhance performance. Following \cite{jing2025stereovideotemporallyconsistent}, we employ the frozen Video Depth Anything (VDA) \cite{chen2025videodepthanythingconsistent} prior in the feature extraction stage, while keeping the correlation and refinement modules unchanged. As shown, BiDAStereo achieves substantial performance gains, highlighting both the flexibility of our approach and its compatibility with depth priors.}

\subsection{Comparison with monocular video methods}
\textcolor{black}{In Tab.~\ref{tab:monocular}, we compare BiDAStereo with representative monocular video depth models \cite{hu2024-DepthCrafter, chen2025videodepthanythingconsistent}. Since these models output relative depth maps, we align them to RAFTStereo \cite{lipson2021raft}'s disparities using a least-squares scale and shift optimization. As shown, our stereo method achieve substantially higher accuracy, outperforming monocular approaches by up to one order of magnitude in EPE. Moreover, BiDAStereo has served as the teacher model for generating pseudo labels during the training of both \cite{hu2024-DepthCrafter, chen2025videodepthanythingconsistent}, further demonstrating the broad applicability and impact of our framework.}

\subsection{Computational Efficiency}
\textcolor{black}{In Tab.~\ref{tab:efficiency}, we compare the methods based on trainable parameters, inference GPU memory, multiply-accumulate operations (MACs) at a resolution of 720$\times$1280, and training iterations. From this table, it is evident that the image-based methods, RAFTStereo and IGEVStereo, exhibit the lowest memory consumption as they process frames independently. In contrast, video-based methods such as DynamicStereo and our BiDAStereo show higher computational costs across all metrics. However, our lightweight BiDAStabilizer demonstrates superior efficiency in terms of MACs and training iterations. Even with multiple frame processing, the memory usage remains within an acceptable range.}

\section{Discussion}

\textcolor{black}{The alignment of different time steps can also be regarded as a Multi-View Stereo (MVS) problem \cite{yao2018mvsnet}, which utilizes explicit geometry information through triangulation. In MVS, the plane sweep algorithm \cite{yao2018mvsnet} is commonly employed to project source images onto fronto-parallel planes of the reference image using homography. However, when dynamic objects are present, the triangulation rule typically applied in MVS becomes inapplicable. Moreover, MVS methods require camera pose information and a pre-defined depth range obtained via Structure from Motion (SfM). Estimating these parameters increases the overall complexity of the pipeline and can introduce additional sources of error.}

\textcolor{black}{Another approach is scene flow \cite{wang2020flownet3d++}, which estimates pixel-wise 3D motions and obviates the need for camera pose estimation. Alignment in this context can be performed using an SE3 transformation, similar to the method \cite{li2023temporally}. However, estimating scene flow requires depth of source images or Lidar information, thereby leading to a chicken-and-egg problem for our task. While iterative optimization techniques can be employed to address this issue, the overall performance remains heavily dependent on the accuracy of the scene flow estimation, as demonstrated in \cite{li2023temporally}.}

\textcolor{black}{In contrast to these explicit methods, optical flow serves as an implicit alignment technique, free from geometric constraints. It establishes correspondences between 3D points in two camera coordinates directly, making it suitable for both dynamic and static objects. In our approaches, optical flow aligns adjacent features and propagates temporal information. Consequently, matching points in different frames should align accurately after adjustment, thereby maintaining temporal consistency.}

\section{Conclusion}
\textcolor{black}{In this paper, we demonstrate that bidirectional alignment is an effective operation that enhances the temporal consistency of video stereo matching. Building on this foundation, we propose a plugin network that can be integrated with existing image-based methods to process video sequences. We introduce a video processing framework, where a triple-frame correlation layer and motion propagation recurrent unit are designed to leverage both local and global temporal information. Regarding the limitations in current datasets, we also develop a synthetic stereo video benchmark and a real-world camera-captured dataset. Our experimental results indicate that the proposed approaches achieve strong performance across a variety of datasets.}

\small 
\bibliographystyle{IEEEbib}
\bibliography{main}

@String(PAMI  = {IEEE Trans. Pattern Anal. Mach. Intell.})

@String(IJCV  = {Int. J. Comput. Vis.})

@String(CVPR  = {IEEE Conf. Comput. Vis. Pattern Recog.})

@String(ICCV  = {Int. Conf. Comput. Vis.})

@String(ECCV  = {Eur. Conf. Comput. Vis.})

@String(CVPRW = {IEEE Conf. Comput. Vis. Pattern Recog. Worksh.})

@String(ICPR  = {Int. Conf. Pattern Recog.})

@String(TOG   = {ACM Trans. Graph.})

@String(PAMI  = {IEEE TPAMI})

@String(IJCV  = {IJCV})

@String(CVPR  = {CVPR})

@String(ICCV  = {ICCV})

@String(ECCV  = {ECCV})

@String(CVPRW = {CVPRW})

@String(ICPR  = {ICPR})

@String(TOG   = {ACM TOG})

@inproceedings{mayer2016large,
  title={A large dataset to train convolutional networks for disparity, optical flow, and scene flow estimation},
  author={Mayer, Nikolaus and Ilg, Eddy and Hausser, Philip and Fischer, Philipp and Cremers, Daniel and Dosovitskiy, Alexey and Brox, Thomas},
  booktitle=CVPR,
  pages={4040--4048},
  year={2016}
}

@inproceedings{eth3d,
  title={A multi-view stereo benchmark with high-resolution images and multi-camera videos},
  author={Schops, Thomas and Schonberger, Johannes L and Galliani, Silvano and Sattler, Torsten and Schindler, Konrad and Pollefeys, Marc and Geiger, Andreas},
  booktitle=CVPR,
  pages={3260--3269},
  year={2017}
}

@article{middlebury,
  title={A taxonomy and evaluation of dense two-frame stereo correspondence algorithms},
  author={Scharstein, Daniel and Szeliski, Richard},
  journal=IJCV,
  volume={47},
  number={1},
  pages={7--42},
  year={2002}
}

@inproceedings{kitti,
  title={Are we ready for autonomous driving? the kitti vision benchmark suite},
  author={Geiger, Andreas and Lenz, Philip and Urtasun, Raquel},
  booktitle=CVPR,
  pages={3354--3361},
  year={2012}
}

@inproceedings{sintel,
title = {A naturalistic open source movie for optical flow evaluation},
author = {Butler, D. J. and Wulff, J. and Stanley, G. B. and Black, M. J.},
booktitle = ECCV,
pages = {611--625},
year = {2012}
}

@inproceedings{fallingthings,
  title={Falling things: A synthetic dataset for 3d object detection and pose estimation},
  author={Tremblay, Jonathan and To, Thang and Birchfield, Stan},
  booktitle=CVPRW,
  pages={2038--2041},
  year={2018}
}

@Manual{blender,
   title = {Blender - a 3D modelling and rendering package},
   author = {{Blender Online Community}},
   organization = {Blender Foundation},
   address = {Blender Institute, Amsterdam},
   year = {2021}
}

@misc{skorobogatova2022real,
  title={Real-Time Global Illumination in Unreal Engine 5},
  author={SKOROBOGATOVA, ANNA},
  year={2022},
  publisher={Masaryk University}
}

@article{tartanair2020iros,
  title =   {TartanAir: A Dataset to Push the Limits of Visual SLAM},
  author =  {Wang, Wenshan and Zhu, Delong and Wang, Xiangwei and Hu, Yaoyu and Qiu, Yuheng and Wang, Chen and Hu, Yafei and Kapoor, Ashish and Scherer, Sebastian},
  booktitle = {2020 IEEE/RSJ International Conference on Intelligent Robots and Systems (IROS)},
  year =    {2020}
}

@inproceedings{kittidepth,
  title={Sparsity invariant cnns},
  author={Uhrig, Jonas and Schneider, Nick and Schneider, Lukas and Franke, Uwe and Brox, Thomas and Geiger, Andreas},
  booktitle={2017 international conference on 3D Vision (3DV)},
  pages={11--20},
  year={2017},
  organization={IEEE}
}

@inproceedings{yang2019drivingstereo,
  title={Drivingstereo: A large-scale dataset for stereo matching in autonomous driving scenarios},
  author={Yang, Guorun and Song, Xiao and Huang, Chaoqin and Deng, Zhidong and Shi, Jianping and Zhou, Bolei},
  booktitle={Proceedings of the IEEE/CVF Conference on Computer Vision and Pattern Recognition},
  pages={899--908},
  year={2019}
}

@article{birchfield1999depth,
  title={Depth discontinuities by pixel-to-pixel stereo},
  author={Birchfield, Stan and Tomasi, Carlo},
  journal=IJCV,
  volume={35},
  number={3},
  pages={269--293},
  year={1999}
}

@article{van2002hierarchical,
  title={A hierarchical symmetric stereo algorithm using dynamic programming},
  author={Van Meerbergen, Geert and Vergauwen, Maarten and Pollefeys, Marc and Van Gool, Luc},
  journal=IJCV,
  volume={47},
  number={1},
  pages={275--285},
  year={2002}
}

@article{scharstein2002taxonomy,
  title={A taxonomy and evaluation of dense two-frame stereo correspondence algorithms},
  author={Scharstein, Daniel and Szeliski, Richard},
  journal=IJCV,
  volume={47},
  number={1},
  pages={7--42},
  year={2002}
}

@article{boykov2001fast,
  title={Fast approximate energy minimization via graph cuts},
  author={Boykov, Yuri and Veksler, Olga and Zabih, Ramin},
  journal=PAMI,
  volume={23},
  number={11},
  pages={1222--1239},
  year={2001}
}

@article{sun2003stereo,
  title={Stereo matching using belief propagation},
  author={Sun, Jian and Zheng, Nan-Ning and Shum, Heung-Yeung},
  journal=PAMI,
  volume={25},
  number={7},
  pages={787--800},
  year={2003}
}

@inproceedings{klaus2006segment,
  title={Segment-based stereo matching using belief propagation and a self-adapting dissimilarity measure},
  author={Klaus, Andreas and Sormann, Mario and Karner, Konrad},
  booktitle=ICPR,
  volume={3},
  pages={15--18},
  year={2006}
}

@article{kanade1994stereo,
  title={A stereo matching algorithm with an adaptive window: Theory and experiment},
  author={Kanade, Takeo and Okutomi, Masatoshi},
  journal={IEEE transactions on pattern analysis and machine intelligence},
  volume={16},
  number={9},
  pages={920--932},
  year={1994},
  publisher={IEEE}
}

@article{scharstein1998stereo,
  title={Stereo matching with nonlinear diffusion},
  author={Scharstein, Daniel and Szeliski, Richard},
  journal={International journal of computer vision},
  volume={28},
  pages={155--174},
  year={1998},
  publisher={Springer}
}

@article{hsieh1992performance,
  title={Performance evaluation of scene registration and stereo matching for artographic feature extraction},
  author={Hsieh, Yuan C and McKeown, David M and Perlant, Frederic P},
  journal={IEEE Transactions on Pattern Analysis \& Machine Intelligence},
  volume={14},
  number={02},
  pages={214--238},
  year={1992},
  publisher={IEEE Computer Society}
}

@article{jenkin1986applying,
  title={Applying temporal constraints to the dynamic stereo problem},
  author={Jenkin, Michael and Tsotsos, John K},
  journal={Computer Vision, Graphics, and Image Processing},
  volume={33},
  number={1},
  pages={16--32},
  year={1986},
  publisher={Elsevier}
}

@article{sand2004video,
  title={Video matching},
  author={Sand, Peter and Teller, Seth},
  journal={Acm transactions on graphics (tog)},
  volume={23},
  number={3},
  pages={592--599},
  year={2004},
  publisher={ACM New York, NY, USA}
}

@inproceedings{min2010temporally,
  title={Temporally consistent stereo matching using coherence function},
  author={Min, Dongbo and Yea, Sehoon and Vetro, Anthony},
  booktitle={2010 3DTV-Conference: The True Vision-Capture, Transmission and Display of 3D Video},
  pages={1--4},
  year={2010},
  organization={IEEE}
}

@inproceedings{zbontar2015computing,
  title={Computing the stereo matching cost with a convolutional neural network},
  author={Zbontar, Jure and LeCun, Yann},
  booktitle=CVPR,
  pages={1592--1599},
  year={2015}
}

@inproceedings{pang2017cascade,
  title={Cascade residual learning: A two-stage convolutional neural network for stereo matching},
  author={Pang, Jiahao and Sun, Wenxiu and Ren, Jimmy SJ and Yang, Chengxi and Yan, Qiong},
  booktitle=CVPRW,
  pages={887--895},
  year={2017}
}

@inproceedings{guo2019group,
  title={Group-wise correlation stereo network},
  author={Guo, Xiaoyang and Yang, Kai and Yang, Wukui and Wang, Xiaogang and Li, Hongsheng},
  booktitle=CVPR,
  pages={3273--3282},
  year={2019}
}

@inproceedings{xu2020aanet,
  title={Aanet: Adaptive aggregation network for efficient stereo matching},
  author={Xu, Haofei and Zhang, Juyong},
  booktitle=CVPR,
  pages={1959--1968},
  year={2020}
}

@inproceedings{tankovich2021hitnet,
  title={Hitnet: Hierarchical iterative tile refinement network for real-time stereo matching},
  author={Tankovich, Vladimir and Hane, Christian and Zhang, Yinda and Kowdle, Adarsh and Fanello, Sean and Bouaziz, Sofien},
  booktitle=CVPR,
  pages={14362--14372},
  year={2021}
}

@inproceedings{teed2020raft,
  title={Raft: Recurrent all-pairs field transforms for optical flow},
  author={Teed, Zachary and Deng, Jia},
  booktitle=ECCV,
  pages={402--419},
  year={2020}
}

@article{lipson2021raft,
  title={RAFT-Stereo: Multilevel Recurrent Field Transforms for Stereo Matching},
  author={Lipson, Lahav and Teed, Zachary and Deng, Jia},
  journal={arXiv preprint arXiv:2109.07547},
  year={2021}
}

@inproceedings{song2021adastereo,
  title={AdaStereo: a simple and efficient approach for adaptive stereo matching},
  author={Song, Xiao and Yang, Guorun and Zhu, Xinge and Zhou, Hui and Wang, Zhe and Shi, Jianping},
  booktitle=CVPR,
  pages={10328--10337},
  year={2021}
}

@inproceedings{shen2021cfnet,
  title={CFNet: Cascade and Fused Cost Volume for Robust Stereo Matching},
  author={Shen, Zhelun and Dai, Yuchao and Rao, Zhibo},
  booktitle=CVPR,
  pages={13906--13915},
  year={2021}
}

@inproceedings{li2022practical,
  title={Practical stereo matching via cascaded recurrent network with adaptive correlation},
  author={Li, Jiankun and Wang, Peisen and Xiong, Pengfei and Cai, Tao and Yan, Ziwei and Yang, Lei and Liu, Jiangyu and Fan, Haoqiang and Liu, Shuaicheng},
  booktitle={Proceedings of the IEEE/CVF Conference on Computer Vision and Pattern Recognition},
  pages={16263--16272},
  year={2022}
}

@inproceedings{kendall2017end,
  title={End-to-end learning of geometry and context for deep stereo regression},
  author={Kendall, Alex and Martirosyan, Hayk and Dasgupta, Saumitro and Henry, Peter and Kennedy, Ryan and Bachrach, Abraham and Bry, Adam},
  booktitle=CVPR,
  pages={66--75},
  year={2017}
}

@inproceedings{chang2018pyramid,
  title={Pyramid stereo matching network},
  author={Chang, Jia-Ren and Chen, Yong-Sheng},
  booktitle=CVPR,
  pages={5410--5418},
  year={2018}
}

@inproceedings{zhang2019ga,
  title={Ga-net: Guided aggregation net for end-to-end stereo matching},
  author={Zhang, Feihu and Prisacariu, Victor and Yang, Ruigang and Torr, Philip HS},
  booktitle=CVPR,
  pages={185--194},
  year={2019}
}

@inproceedings{yang2019hierarchical,
  title={Hierarchical deep stereo matching on high-resolution images},
  author={Yang, Gengshan and Manela, Joshua and Happold, Michael and Ramanan, Deva},
  booktitle=CVPR,
  pages={5515--5524},
  year={2019}
}

@inproceedings{pang2018zoom,
  title={Zoom and learn: Generalizing deep stereo matching to novel domains},
  author={Pang, Jiahao and Sun, Wenxiu and Yang, Chengxi and Ren, Jimmy and Xiao, Ruichao and Zeng, Jin and Lin, Liang},
  booktitle=CVPR,
  pages={2070--2079},
  year={2018}
}

@article{adam,
  title={Adam: A method for stochastic optimization},
  author={Kingma, Diederik P and Ba, Jimmy},
  journal={arXiv preprint arXiv:1412.6980},
  year={2014}
}

@article{xu2022acvnet,
  title={Accurate and Efficient Stereo Matching via Attention Concatenation Volume},
  author={Xu, Gangwei and Wang, Yun and Cheng, Junda and Tang, Jinhui and Yang, Xin},
  journal={arXiv preprint arXiv:2209.12699},
  year={2022}
}

@inproceedings{geiger2011stereoscan,
  title={Stereoscan: Dense 3d reconstruction in real-time},
  author={Geiger, Andreas and Ziegler, Julius and Stiller, Christoph},
  booktitle={2011 IEEE intelligent vehicles symposium (IV)},
  pages={963--968},
  year={2011},
  organization={Ieee}
}

@inproceedings{xu2023iterative,
  title={Iterative Geometry Encoding Volume for Stereo Matching},
  author={Xu, Gangwei and Wang, Xianqi and Ding, Xiaohuan and Yang, Xin},
  booktitle={Proceedings of the IEEE/CVF Conference on Computer Vision and Pattern Recognition},
  pages={21919--21928},
  year={2023}
}

@inproceedings{li2023temporally,
  title={Temporally consistent online depth estimation in dynamic scenes},
  author={Li, Zhaoshuo and Ye, Wei and Wang, Dilin and Creighton, Francis X and Taylor, Russell H and Venkatesh, Ganesh and Unberath, Mathias},
  booktitle={Proceedings of the IEEE/CVF winter conference on applications of computer vision},
  pages={3018--3027},
  year={2023}
}

@inproceedings{zhang2023temporalstereo,
  title={Temporalstereo: Efficient spatial-temporal stereo matching network},
  author={Zhang, Youmin and Poggi, Matteo and Mattoccia, Stefano},
  booktitle={2023 IEEE/RSJ International Conference on Intelligent Robots and Systems (IROS)},
  pages={9528--9535},
  year={2023},
  organization={IEEE}
}

@article{reis2023real,
  title={Real-time flying object detection with YOLOv8},
  author={Reis, Dillon and Kupec, Jordan and Hong, Jacqueline and Daoudi, Ahmad},
  journal={arXiv preprint arXiv:2305.09972},
  year={2023}
}

@inproceedings{cheng2024stereo,
  title={Stereo Matching in Time: 100+ FPS Video Stereo Matching for Extended Reality},
  author={Cheng, Ziang and Yang, Jiayu and Li, Hongdong},
  booktitle={Proceedings of the IEEE/CVF Winter Conference on Applications of Computer Vision},
  pages={8719--8728},
  year={2024}
}

@inproceedings{karaev2023dynamicstereo,
  title={DynamicStereo: Consistent Dynamic Depth from Stereo Videos},
  author={Karaev, Nikita and Rocco, Ignacio and Graham, Benjamin and Neverova, Natalia and Vedaldi, Andrea and Rupprecht, Christian},
  booktitle={Proceedings of the IEEE/CVF Conference on Computer Vision and Pattern Recognition},
  pages={13229--13239},
  year={2023}
}

@inproceedings{khan2023temporally,
  title={Temporally consistent online depth estimation using point-based fusion},
  author={Khan, Numair and Penner, Eric and Lanman, Douglas and Xiao, Lei},
  booktitle={Proceedings of the IEEE/CVF Conference on Computer Vision and Pattern Recognition},
  pages={9119--9129},
  year={2023}
}

@inproceedings{zhong2018open,
  title={Open-world stereo video matching with deep rnn},
  author={Zhong, Yiran and Li, Hongdong and Dai, Yuchao},
  booktitle={Proceedings of the European Conference on Computer Vision (ECCV)},
  pages={101--116},
  year={2018}
}

@inproceedings{smith2019super,
  title={Super-convergence: Very fast training of neural networks using large learning rates},
  author={Smith, Leslie N and Topin, Nicholay},
  booktitle={Artificial intelligence and machine learning for multi-domain operations applications},
  volume={11006},
  pages={369--386},
  year={2019},
  organization={SPIE}
}

@inproceedings{rao2023masked,
  title={Masked representation learning for domain generalized stereo matching},
  author={Rao, Zhibo and Xiong, Bangshu and He, Mingyi and Dai, Yuchao and He, Renjie and Shen, Zhelun and Li, Xing},
  booktitle={Proceedings of the IEEE/CVF Conference on Computer Vision and Pattern Recognition},
  pages={5435--5444},
  year={2023}
}

@inproceedings{chang2023domain,
  title={Domain Generalized Stereo Matching via Hierarchical Visual Transformation},
  author={Chang, Tianyu and Yang, Xun and Zhang, Tianzhu and Wang, Meng},
  booktitle={Proceedings of the IEEE/CVF Conference on Computer Vision and Pattern Recognition},
  pages={9559--9568},
  year={2023}
}

@inproceedings{chuah2022itsa,
  title={Itsa: An information-theoretic approach to automatic shortcut avoidance and domain generalization in stereo matching networks},
  author={Chuah, WeiQin and Tennakoon, Ruwan and Hoseinnezhad, Reza and Bab-Hadiashar, Alireza and Suter, David},
  booktitle={Proceedings of the IEEE/CVF Conference on Computer Vision and Pattern Recognition},
  pages={13022--13032},
  year={2022}
}

@inproceedings{zhang2022revisiting,
  title={Revisiting domain generalized stereo matching networks from a feature consistency perspective},
  author={Zhang, Jiawei and Wang, Xiang and Bai, Xiao and Wang, Chen and Huang, Lei and Chen, Yimin and Gu, Lin and Zhou, Jun and Harada, Tatsuya and Hancock, Edwin R},
  booktitle={Proceedings of the IEEE/CVF Conference on Computer Vision and Pattern Recognition},
  pages={13001--13011},
  year={2022}
}

@ARTICLE{ucfnet,
  author={Shen, Zhelun and Song, Xibin and Dai, Yuchao and Zhou, Dingfu and Rao, Zhibo and Zhang, Liangjun},
  journal={IEEE Transactions on Pattern Analysis and Machine Intelligence}, 
  title={Digging Into Uncertainty-Based Pseudo-Label for Robust Stereo Matching}, 
  year={2023},
  volume={},
  number={},
  pages={1-18},
  doi={10.1109/TPAMI.2023.3300976}}

@InProceedings{Jing_2023_ICCV,
    author    = {Jing, Junpeng and Li, Jiankun and Xiong, Pengfei and Liu, Jiangyu and Liu, Shuaicheng and Guo, Yichen and Deng, Xin and Xu, Mai and Jiang, Lai and Sigal, Leonid},
    title     = {Uncertainty Guided Adaptive Warping for Robust and Efficient Stereo Matching},
    booktitle = {Proceedings of the IEEE/CVF International Conference on Computer Vision (ICCV)},
    month     = {October},
    year      = {2023},
    pages     = {3318-3327}
}

@article{desouza2002vision,
  title={Vision for mobile robot navigation: A survey},
  author={DeSouza, Guilherme N and Kak, Avinash C},
  journal={IEEE transactions on pattern analysis and machine intelligence},
  volume={24},
  number={2},
  pages={237--267},
  year={2002},
  publisher={IEEE}
}

@article{azuma1997survey,
  title={A survey of augmented reality},
  author={Azuma, Ronald T},
  journal={Presence: teleoperators \& virtual environments},
  volume={6},
  number={4},
  pages={355--385},
  year={1997},
  publisher={MIT Press One Rogers Street, Cambridge, MA 02142-1209, USA journals-info~…}
}

@book{hartley2003multiple,
  title={Multiple view geometry in computer vision},
  author={Hartley, Richard and Zisserman, Andrew},
  year={2003},
  publisher={Cambridge university press}
}

@article{jing2024matchstereovideos,
  title={Match-Stereo-Videos: Bidirectional Alignment for Consistent Dynamic Stereo Matching}, 
  author={Junpeng Jing and Ye Mao and Krystian Mikolajczyk},
  year={2024},
  eprint={2403.10755},
  archivePrefix={arXiv},
  primaryClass={cs.CV}
  }

@inproceedings{wang2020flownet3d++,
  title={Flownet3d++: Geometric losses for deep scene flow estimation},
  author={Wang, Zirui and Li, Shuda and Howard-Jenkins, Henry and Prisacariu, Victor and Chen, Min},
  booktitle={Proceedings of the IEEE/CVF winter conference on applications of computer vision},
  pages={91--98},
  year={2020}
}

@inproceedings{yao2018mvsnet,
  title={Mvsnet: Depth inference for unstructured multi-view stereo},
  author={Yao, Yao and Luo, Zixin and Li, Shiwei and Fang, Tian and Quan, Long},
  booktitle={Proceedings of the European conference on computer vision (ECCV)},
  pages={767--783},
  year={2018}
}

@inproceedings{kopf2021robust,
  title={Robust consistent video depth estimation},
  author={Kopf, Johannes and Rong, Xuejian and Huang, Jia-Bin},
  booktitle={Proceedings of the IEEE/CVF Conference on Computer Vision and Pattern Recognition},
  pages={1611--1621},
  year={2021}
}

@article{luo2020consistent,
  title={Consistent video depth estimation},
  author={Luo, Xuan and Huang, Jia-Bin and Szeliski, Richard and Matzen, Kevin and Kopf, Johannes},
  journal={ACM Transactions on Graphics (ToG)},
  volume={39},
  number={4},
  pages={71--1},
  year={2020},
  publisher={ACM New York, NY, USA}
}

@inproceedings{li2023towards,
  title={Towards practical consistent video depth estimation},
  author={Li, Pengzhi and Ding, Yikang and Li, Linge and Guan, Jingwei and Li, Zhiheng},
  booktitle={Proceedings of the 2023 ACM International Conference on Multimedia Retrieval},
  pages={388--397},
  year={2023}
}

@article{hu2024-DepthCrafter,
            author      = {Hu, Wenbo and Gao, Xiangjun and Li, Xiaoyu and Zhao, Sijie and Cun, Xiaodong and Zhang, Yong and Quan, Long and Shan, Ying},
            title       = {DepthCrafter: Generating Consistent Long Depth Sequences for Open-world Videos},
            journal     = {arXiv preprint arXiv:2409.02095},
            year        = {2024}
    }

@misc{chen2025videodepthanythingconsistent,
      title={Video Depth Anything: Consistent Depth Estimation for Super-Long Videos}, 
      author={Sili Chen and Hengkai Guo and Shengnan Zhu and Feihu Zhang and Zilong Huang and Jiashi Feng and Bingyi Kang},
      year={2025},
      eprint={2501.12375},
      archivePrefix={arXiv},
      primaryClass={cs.CV},
      url={https://arxiv.org/abs/2501.12375}, 
}

@misc{jing2025stereovideotemporallyconsistent,
                        title={Stereo Any Video: Temporally Consistent Stereo Matching}, 
                        author={Junpeng Jing and Weixun Luo and Ye Mao and Krystian Mikolajczyk},
                        year={2025},
                        eprint={2503.05549},
                        archivePrefix={arXiv},
                        primaryClass={cs.CV},
                        url={https://arxiv.org/abs/2503.05549}, 
                }

@inproceedings{cheng2022masked,
  title={Masked-attention mask transformer for universal image segmentation},
  author={Cheng, Bowen and Misra, Ishan and Schwing, Alexander G and Kirillov, Alexander and Girdhar, Rohit},
  booktitle={Proceedings of the IEEE/CVF conference on computer vision and pattern recognition},
  pages={1290--1299},
  year={2022}
}

@inproceedings{raistrick2023infinite,
  title={Infinite photorealistic worlds using procedural generation},
  author={Raistrick, Alexander and Lipson, Lahav and Ma, Zeyu and Mei, Lingjie and Wang, Mingzhe and Zuo, Yiming and Kayan, Karhan and Wen, Hongyu and Han, Beining and Wang, Yihan and others},
  booktitle={Proceedings of the IEEE/CVF conference on computer vision and pattern recognition},
  pages={12630--12641},
  year={2023}
}

@inproceedings{cao2021learning,
  title={Learning structure affinity for video depth estimation},
  author={Cao, Yuanzhouhan and Li, Yidong and Zhang, Haokui and Ren, Chao and Liu, Yifan},
  booktitle={Proceedings of the 29th ACM International Conference on Multimedia},
  pages={190--198},
  year={2021}
}

@inproceedings{lai2018learning,
  title={Learning blind video temporal consistency},
  author={Lai, Wei-Sheng and Huang, Jia-Bin and Wang, Oliver and Shechtman, Eli and Yumer, Ersin and Yang, Ming-Hsuan},
  booktitle={Proceedings of the European conference on computer vision (ECCV)},
  pages={170--185},
  year={2018}
}

@inproceedings{wang2022less,
  title={Less is more: Consistent video depth estimation with masked frames modeling},
  author={Wang, Yiran and Pan, Zhiyu and Li, Xingyi and Cao, Zhiguo and Xian, Ke and Zhang, Jianming},
  booktitle={Proceedings of the 30th ACM International Conference on Multimedia},
  pages={6347--6358},
  year={2022}
}

@inproceedings{li2021enforcing,
  title={Enforcing temporal consistency in video depth estimation},
  author={Li, Siyuan and Luo, Yue and Zhu, Ye and Zhao, Xun and Li, Yu and Shan, Ying},
  booktitle={Proceedings of the IEEE/CVF International Conference on Computer Vision},
  pages={1145--1154},
  year={2021}
}

@article{wang2004image,
  title={Image quality assessment: from error visibility to structural similarity},
  author={Wang, Zhou and Bovik, Alan C and Sheikh, Hamid R and Simoncelli, Eero P},
  journal={IEEE transactions on image processing},
  volume={13},
  number={4},
  pages={600--612},
  year={2004},
  publisher={IEEE}
}

@InProceedings{Wang_2023_ICCV,
    author    = {Wang, Yiran and Shi, Min and Li, Jiaqi and Huang, Zihao and Cao, Zhiguo and Zhang, Jianming and Xian, Ke and Lin, Guosheng},
    title     = {Neural Video Depth Stabilizer},
    booktitle = {Proceedings of the IEEE/CVF International Conference on Computer Vision (ICCV)},
    month     = {October},
    year      = {2023},
    pages     = {9466-9476}
}

@misc{robustvision,
  author = {{Robust Vision Challenge Organizing Team}},
  title = {Robust Vision Challenge Website},
  howpublished = {\url{http://www.robustvision.net/index.php}},
  note = {Accessed: 2024-09-18}
}

\begin{IEEEbiography}[{\includegraphics[width=1in,height=1.25in,clip,keepaspectratio]{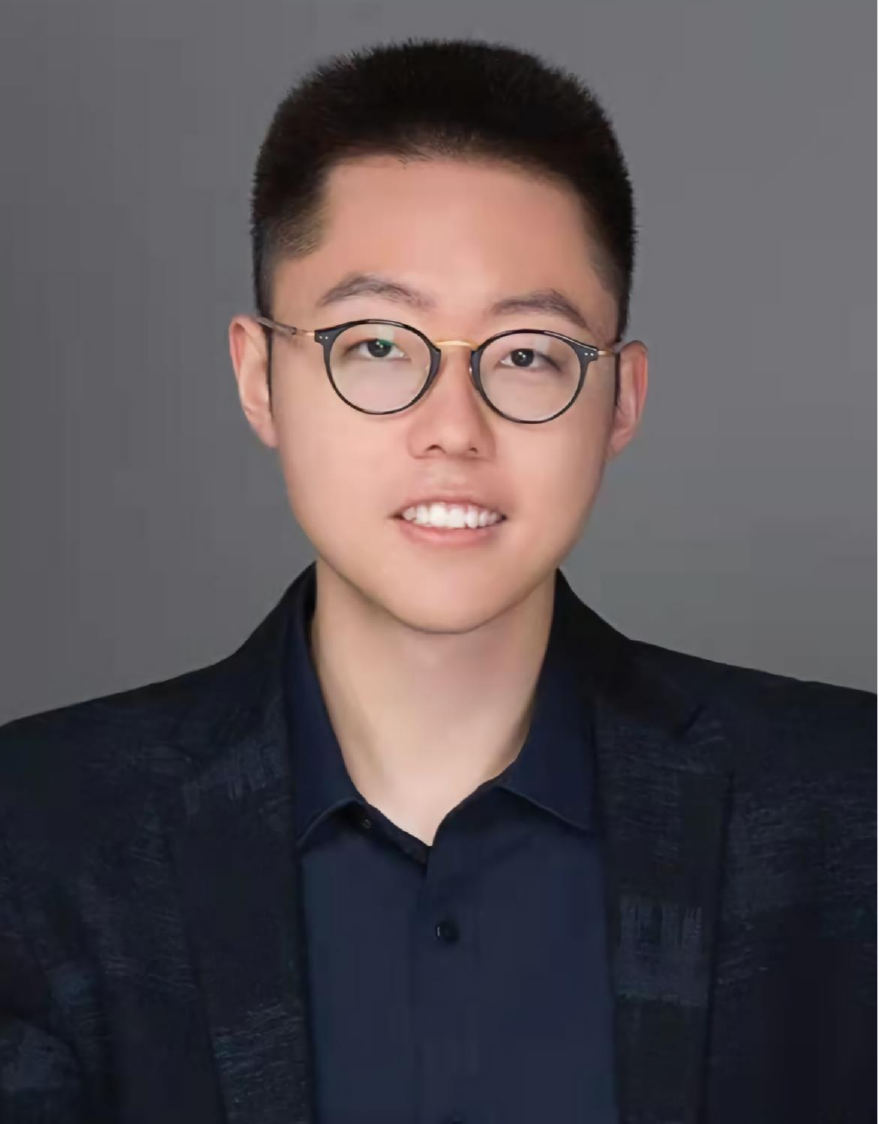}}]{Junpeng Jing}
received the Ph.D. degree from Imperial College London, U.K., in 2026. He received the B.Eng. and M.Eng. degrees from Beihang University, China, in 2020 and 2023, respectively. His research interests include stereo depth estimation, 3D learning and understanding, and low-level vision. In 2023, he won the champion of the Robust Vision Challenge. He has published papers in top-tier journals and conferences, including IEEE TPAMI, CVPR, ICCV, ECCV, NeurIPS, and ICML, with several recognized as ESI highly cited and highlight papers.
\end{IEEEbiography}
\vfill

\begin{IEEEbiography}[{\includegraphics[width=1in,height=1.25in,clip,keepaspectratio]{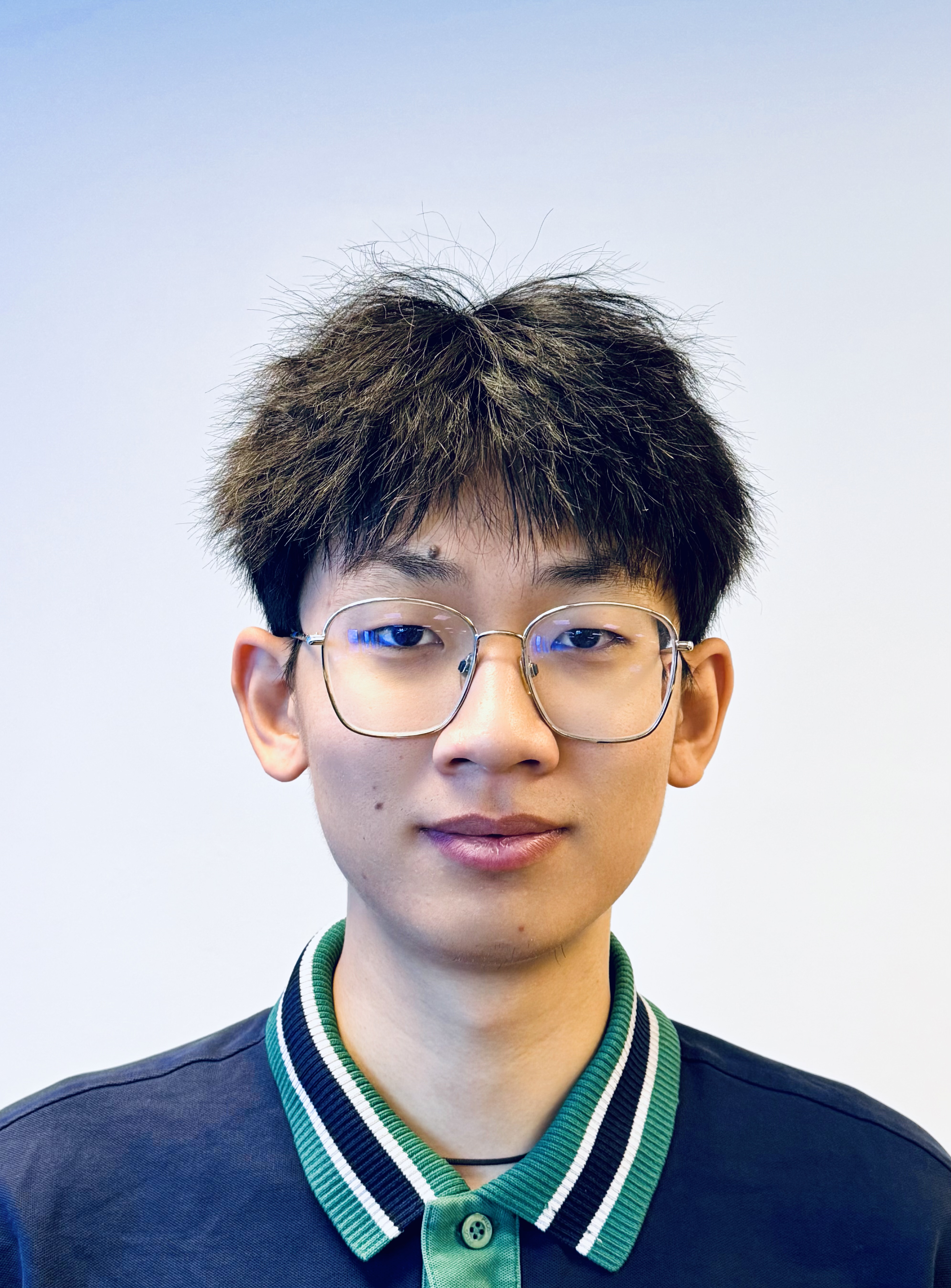}}]{Ye Mao} is a PhD candidate at Imperial College London, where the Imperial President's Scholarship funds him. He received his BSc degree in computer science from King's College London in 2021 and completed his MPhil at the University of Cambridge in 2023. His research primarily focuses on 3D computer vision and vision-language models. Ye has authored several papers published in top-tier journals and conferences, including IEEE JBHI, ICCV, ECCV, NeurIPS, ICML and MICCAI.
\end{IEEEbiography}
\vfill

\begin{IEEEbiography}[{\includegraphics[width=1in,height=1.25in,clip,keepaspectratio]{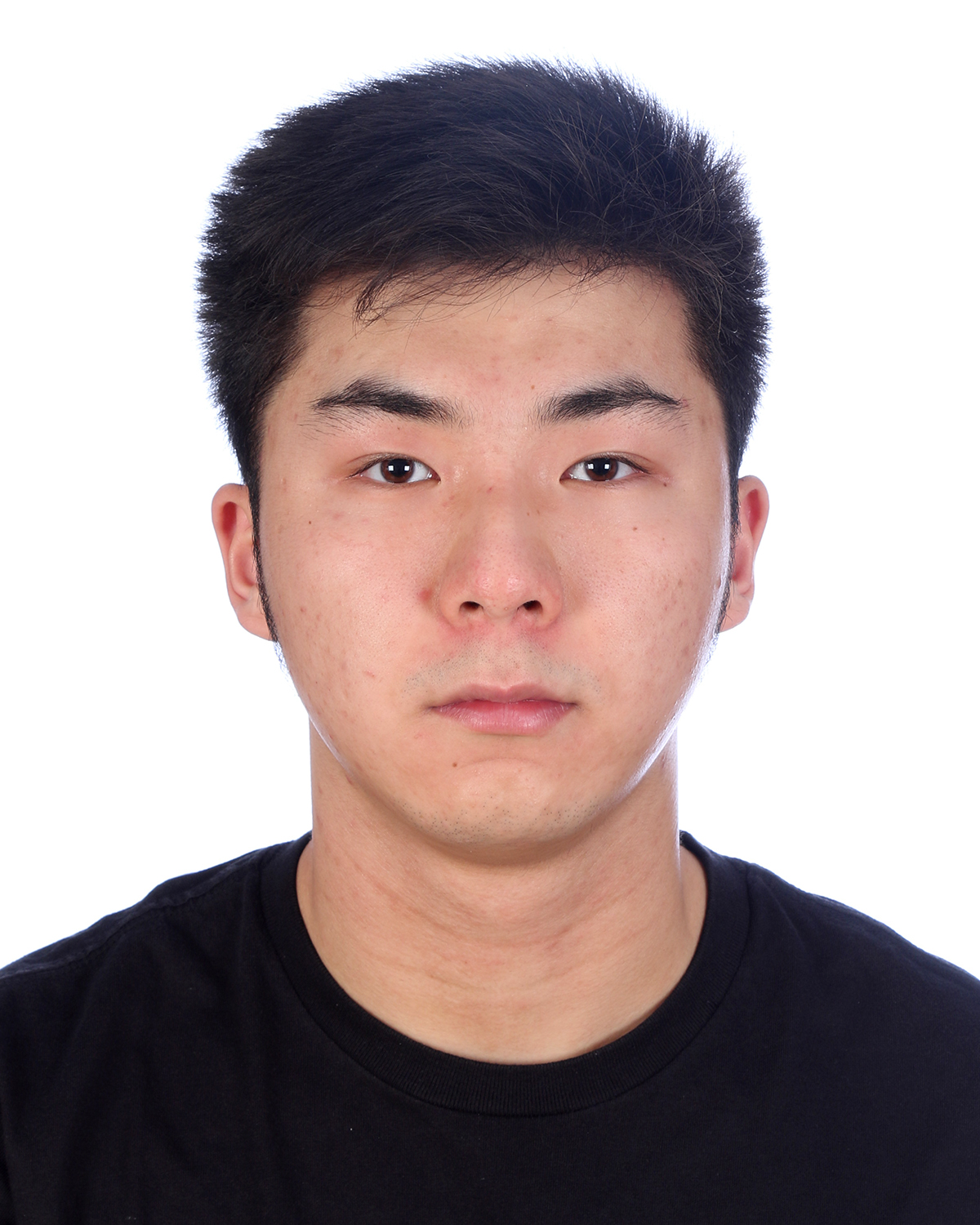}}]{Anlan Qiu} received his M.Eng. degree in Electronic and Information Engineering from Imperial College London in 2025. His research interests include machine learning and computer vision.
\end{IEEEbiography}
\vfill

\begin{IEEEbiography}[{\includegraphics[width=1in,height=1.25in,clip,keepaspectratio]{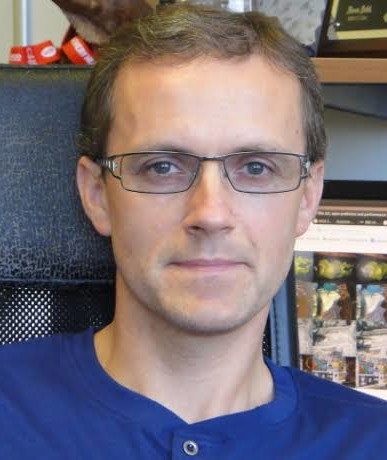}}]{Krystian Mikolajczyk} received the PhD degree from the Institute National Polytechnique de Grenoble and held a number of research positions at INRIA, University of Oxford and Technical University of Darmstadt, as well as faculty positions at the University of Surrey, and Imperial College London. He is a professor at Imperial College London. His main area of expertise is in image and video recognition, in particular methods for image representation and learning. He has served in various roles at major international conferences co-chairing British Machine Vision Conference 2012 and IEEE International Conference on Advanced Video and Signal-Based Surveillance 2013. In 2014 he received Longuet-Higgins Prize awarded by the Technical Committee on Pattern Analysis and Machine Intelligence of the IEEE Computer Society.

\end{IEEEbiography}
\vfill

\end{document}